\documentclass[10pt,twocolumn,letterpaper]{article}

\usepackage{cvpr}              

\usepackage[dvipsnames]{xcolor}

\usepackage{float}
\usepackage{algorithm}
\usepackage{algpseudocode}
\usepackage{makecell}
\usepackage{multirow}

\usepackage[T1]{fontenc}
\usepackage[font=small,labelfont=bf,tableposition=top]{caption}

\newcommand{\alg}{FedDGM}

\definecolor{cvprblue}{rgb}{0.21,0.49,0.74}
\usepackage[pagebackref,breaklinks,colorlinks,citecolor=cvprblue]{hyperref}

\def\paperID{16631} 
\def\confName{CVPR}
\def\confYear{2024}

\title{ Unlocking the Potential of Federated Learning: The Symphony of Dataset Distillation via Deep Generative Latents}

\author{Yuqi Jia$^{*1}\quad$Saeed Vahidian$^{*1}\quad$Jingwei Sun$^{1}\quad$\\
Jianyi Zhang$^{1}\quad $Vyacheslav Kungurtsev$^{2} \quad$Neil Zhenqiang Gong$^{1}\quad$Yiran Chen$^{1}\quad$\\
$^{1}$Duke University $\quad^{2}$ Czech Technical University\\
{\tt\small saeed.vahidian@duke.edu}\\
{$^*$\small Equal contribution}
}

\begin{document}
\maketitle

\begin{abstract}
\vspace{-2mm}
Data heterogeneity presents significant challenges for federated learning (FL). Recently, dataset distillation techniques have been introduced, and performed at the client level, to attempt to mitigate some of these challenges. In this paper, we propose a highly efficient FL dataset distillation framework on the \emph{server} side, significantly reducing both the computational and communication demands on local devices while enhancing the clients' privacy. Unlike previous strategies that perform dataset distillation on local devices and upload synthetic data to the server, our technique enables the server to leverage prior knowledge from pre-trained deep generative models to synthesize essential data representations from a heterogeneous model architecture. This process allows local devices to train smaller surrogate models while enabling the training of a larger global model on the server, effectively minimizing resource utilization. We substantiate our claim with a theoretical analysis, demonstrating the asymptotic resemblance of the process to the hypothetical ideal of completely centralized training on a heterogeneous dataset. Empirical evidence from our comprehensive experiments indicates our method's superiority, delivering an accuracy enhancement of up to 40\% over non-dataset-distillation techniques in highly heterogeneous FL contexts, and surpassing existing dataset-distillation methods by 18\%.  In addition to the high accuracy, our framework converges faster than the baselines because rather than the server trains on several sets of heterogeneous data distributions, it trains on a multi-modal distribution.  Our code is available at~\url{https://github.com/FedDG23/FedDG-main.git}

\end{abstract}

\vspace{-7mm}
\section{Introduction}



Federated Learning (FL), a recently very popular approach in the realm of IoT and AI, enables a multitude of devices to collaboratively learn a shared model, while keeping the training data localized, thereby addressing privacy concerns inherent in traditional centralized training methods~\cite{Vahidian-ICDCS}. In practice, however, FL faces a significant challenge in the form of data heterogeneity due to the diverse and non-i.i.d. nature of client data, shaped by varying user preferences and usage patterns~\cite{ghosh2020efficient, vahidian-pacfl-2022, Vahidian-CEFHRI}. To address this complex challenge, which impacts the effectiveness and efficiency of the learning process, several previous methods have been proposed, such as \cite{li2018federated, vahidian-curr-FL}. Recent research, such as FedDM~\cite{xiong2022feddm}, integrates dataset distillation (DD) techniques into FL, performing DD on local devices and uploading synthetic data to the server, which has shown exciting progress in addressing the challenge of data heterogeneity.

However, existing methods incorporating DD in FL are not without their limitations. One significant issue is their inability to enhance knowledge generalization across different model architectures, a challenge especially relevant in FL due to the variability of device resources and capabilities. Furthermore, these methods often compromise data privacy principles, as they require clients to upload synthetic data directly to the server. Addressing these concerns, we propose an advanced, server-centric FL DD framework, which significantly bolsters data privacy and substantially enhances knowledge generalization across different model architectures in various settings to reduce the computational load and communication overhead on client devices.  Our methodology is structured into three pivotal components in each communication round. Firstly, it facilitates the training of compact surrogate models on local devices, which are then updated to the server, accommodating diverse resource constraints. Secondly, leveraging pre-trained deep generative models, the server synthesizes distilled data representations via matching training trajectories of the local surrogate models. Subsequently, the server employs this distilled synthetic data to refine a more comprehensive global model. Our extensive experimental analysis underscores the effectiveness of our approach. We demonstrate a notable 40\% increase in accuracy compared to traditional non-dataset-distillation techniques within varied FL environments, as exemplified by our results on the CIFAR-10 benchmark. Furthermore, our method outperforms existing DD techniques by 18\% and shows a remarkable performance improvement of around 10\% on high-resolution image datasets like ImageNet.

\begin{figure}
\begin{center}\includegraphics[scale=0.75]{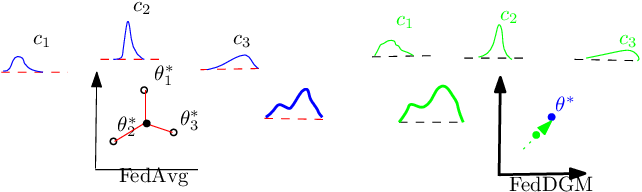}\end{center}
\caption{ Schema of aggregation methods like FedAvg compared to our framework. The ideal problem would be to train on the central multimodal dataset, including all of the data. FedAvg can be considered as solving each problem individually and averaging the final solution, which is not guaranteed to be closer than a certain distance to the true minimum regardless of the quality of the local training. \alg{} attempts to recreate the dataset in the most relevant way for training, by combining distilled datasets from each client. }\label{fig:schema}
\end{figure}

To motivate the principle, consider Figure~\ref{fig:schema}
While standard aggregation-based FL frameworks such as FedAvg average the gradients/parameters from each local training rounds together, this effectively can be thought of as training on the datasets individually and computing the arithmetic mean of the result. Recall fundamentally that the ideal problem of centrally learning the entire dataset, that in the FL context is impossible due to privacy or technical constraints. By contrast, in our schema, we attempt to solve the ideal problem by attempting to recreate the distribution by combining the datasets from those distilled on training each client's data. Thus, if the distribution, as far as SGD iterates are concerned, is entirely identical, we would recreate the ideal training scenario. Of course, this is not possible in practice, however, fundamentally the quality of the training is only limited to the quality of the dataset distillation.

The contributions of our work are as follows:

\begin{enumerate}
    \item We present a novel FL dataset distillation framework, allowing clients to train smaller models to mitigate computational costs, while the server aggregates this information to train a larger model.

    \item  We conduct a theoretical analysis demonstrating that, from an asymptotic perspective, our method is equivalent to the theoretical ideal of centralized training on a multi-modal distribution, representing the sum of heterogeneous data.

    \item We conduct extensive experiments to underscore the effectiveness of our approach. Our empirical evidence demonstrates a significant improvement in model performance in highly heterogeneous FL contexts, setting a new benchmark for dataset distillation methods in FL.
\end{enumerate}

We test our ideas on two widely adopted network architectures on popular datasets in the FL community (CIFAR10, ImageNet) under a wide range of choices of architecture backbones
and compare them with several global state-of-the-art baselines. We have the following findings:\\
\begin{itemize}
    \item \textbf{F1:} The advantages of our dataset distillation-driven FL approach are accentuated in comparison to aggregation-based FL training methods, particularly when distinct groups of clients manifest substantial differences in the distributions of their local data.

    \item \textbf{F2:} Theoretically, we find interesting implications of the distributional limit from the bilevel structure of the dataset distillation problem. Considering marginals at this limit motivates the approximation to the ideal problem with an approximate dataset.

        \item \textbf{F3:} Our method is architecture-agnostic and significantly enhances state-of-the-art performance on high-resolution images by leveraging prior information from a pre-trained generative model.

    \item \textbf{F4:} Our framework converges significantly faster than the state-of-the-arts. The reason is that our framework aggregates the knowledge from the clients by gathering synthetic data rather than averaging the local model parameters, which preserves the local data information much better.
            
\end{itemize}

\section{Related Work}\label{sec:related-work}

\noindent \textbf{Federated Learning.} Federated learning is a decentralized training paradigm composed of two procedures: local training and global aggregation. Therefore, most existing work focuses on either local training~\cite{vahidian-curr-FL, li_model-contrastive_2021, li_federated_2020} or global aggregation~\cite{wang_tackling_2020, yurochkin_bayesian_2019} to learn a better global model. Dataset distillation has facilitated a range of applications, encompassing continual learning~\cite{Hakan-2021-DD-continual}, neural architecture search~\cite{zhao2021dataset, DD-NAS-2020}, and federated learning \cite{FedSynth-2022, xiong2022feddm, huang2023federated}. In particular, FedSynth~\cite{FedSynth-2022} suggests an approach to upstream communication in FL. Instead of sending the model update, each client learns and transmits a compact synthetic dataset. This dataset can then be used as the training data. FedDM~\cite{xiong2022feddm} involves the creation of synthetic datasets on individual clients to align with the loss landscape of the original data by means of distribution matching. Nonetheless, a privacy concern arises in FedDM, and FedSynth in which the synthesized datasets are shared with the server. The authors in \cite {huang2023federated} recommend employing iterative distribution matching to ensure that clients possess an equal share of balanced local synthetic data.\\

\noindent \textbf{Coreset Selection.} The most relevant existing research to our study pertains to the issue of coreset selection. This problem involves taking a fully labeled dataset and aiming to pick a subset from it so that a model trained on this selected subset can closely match the performance of a model trained on the complete dataset~\cite{Vahidian-joneid-cvpr2020}. This technique has proven successful in various applications, including $K$-median clustering~\cite{har2004coresets}, low-rank approximation~\cite{cohen2017input, Vahidian-IEEE-Access-2022}, spectral approximation~\cite{agarwal2004approximating, Vahidian-SCGIGA, li2013iterative}, and Bayesian inference~\cite{Campbell18_ICML}. Traditional coreset construction methods typically employ importance sampling based on sensitivity scores, which reflect the significance of each data point with respect to the objective function being minimized~\cite{har2004coresets, lucic2017training, cohen2017input}. These methods aim to provide high-confidence solutions. More recently, greedy algorithms, a subset of the Frank-Wolfe algorithm, have been introduced to offer worst-case guarantees for problems like Bayesian inference \cite{Campbell18_ICML}.\\

\noindent \textbf{Dataset Distillation} Wang et al.\cite{Dataset-Distillation-main-Tongzhou} pioneered the idea of dataset distillation, where they advocated representing the model's weights in relation to distilled images and refining them through gradient-based hyperparameter optimization. Following that, various research efforts have notably enhanced the results by incorporating methods such as acquiring soft labels~\cite{softlabel-dd-2021}, intensifying the learning signal through gradient matching~\cite{zhao2021dataset}, utilizing augmentations~\cite{Hakan-2021-DD-continual}, long-range trajectory matching by transferring the knowledge
from pretrained experts~\cite{Cazenavette-DD-2022-trajectory}, learning a generative model to
synthesize training data~\cite{DD-cazenavette2023generalizing}.
\section{Method}

\begin{algorithm}
\caption{The \alg{} Framework }\label{alg:fedgan}

\small
\begin{algorithmic}
\item \hspace{-3mm}
\noindent \colorbox[rgb]{1, 0.95, 1}{
\begin{minipage}{0.94\columnwidth}

\textbf{Input:} $M$ clients indexed by $m$,  participating-client number $K$, communication rounds $T_g$, server global model $f$ with $w_{g}^{(0)}$, server surrogate model $f_s$ with $\theta_g^{(0)}$, pre-trained deep generative model $G_g$.

\end{minipage}
}
\item \hspace{-3mm}
\colorbox[gray]{0.95}{
\begin{minipage}{0.94\columnwidth}
\item  \textbf{Server executes:}
\item     \hspace*{\algorithmicindent} initialize global model $f$ with $w_{g}^{(0)}$
\item     \hspace*{\algorithmicindent} initialize global surrogate model $f_s$ with $\theta_g^{(0)}$
\item     \hspace*{\algorithmicindent} \textbf{for } each round $t=0,1,2,...$ \textbf{do}

\item     \hspace*{\algorithmicindent} \quad $\mathbb{S}_t \leftarrow$ (random set of $K$ clients)

\item     \hspace*{\algorithmicindent} \quad \textbf{for} each client $m\in \mathbb{S}_t$ \textbf{in parallel do}

\item     \hspace*{\algorithmicindent} \quad \quad  broadcast $\theta^{(t)}_g$ to clients

\item     \hspace*{\algorithmicindent} \quad \quad  $\theta_m^{(t+1)} \leftarrow \text{ClientUpdate}(\theta^{(t)}_g, D_m)$
\item     \hspace*{\algorithmicindent} \quad \quad  transmit $\theta_m^{(t+1)}$ to the server
\item     \hspace*{\algorithmicindent} \quad \textbf{for} each client $m\in \mathbb{S}_t$ \textbf{do}
\item     \hspace*{\algorithmicindent} \quad \quad {randomly initialize vectors $\tilde{W}_m^{(t)} $}
\item     \hspace*{\algorithmicindent} \quad \quad calculate latent vectors $\tilde{Z}_m^{(t)}$ using $\tilde{W}_m^{(t)}$

\item     \hspace*{\algorithmicindent} \quad \quad \textbf{for} $t_d$ = 1,...,$T_d$ \textbf{do}
\item     \hspace*{\algorithmicindent} \quad \quad \quad generate distilled data $\tilde{D}_m^{(t)}=G_g(\tilde{Z}_m^{(t)})$
\item     \hspace*{\algorithmicindent} \quad \quad \quad initialize student network: $\hat{\theta}_g^{(t)}=\theta_g^{(t)}$
\item     \hspace*{\algorithmicindent} \quad \quad \quad \textbf{for} $t_s=1,2,...T_s$ \textbf{do}
\item     \hspace*{\algorithmicindent} \quad \quad \quad \quad update $\hat{\theta}_g^{(t+t_s)}$ on $\tilde{D}_m^{(t)}$ from $\hat{\theta}_g^{(t+t_s-1)}$ by SGD
\item     \hspace*{\algorithmicindent} \quad \quad \quad compute the Trajectory Matching loss $\mathcal{L}_{MTT}$

\item     \hspace*{\algorithmicindent} \quad \quad \quad update $\tilde{Z}_m^{(t)}$ with respect to $\mathcal{L}_{MTT}$
\item     \hspace*{\algorithmicindent} \quad $\tilde{D}^{(t+1)}_g = \{\tilde{D}_m^{(t)}|\tilde{D}_m^{(t)}=G_g(\tilde{Z}_m^{(t)}), m\in \mathbb{S}_t\}$
\item     \hspace*{\algorithmicindent} \quad update parameters to $w_{g}^{(t+1)}$ on $\tilde{D}^{(t+1)}_g$ by SGD
\item     \hspace*{\algorithmicindent} \quad update parameters to $\theta_g^{(t+1)}$ on $\tilde{D}^{(t+1)}_g$ by SGD
\item     \hspace*{\algorithmicindent} \textbf{return} $\theta_g^{(t+1)}$
\end{minipage}
}
\item \hspace{-3mm}
\colorbox[rgb]{0.95, 0.98, 1}{
\begin{minipage}{0.94\columnwidth}

\item  \textbf{ClientUpdate ($\theta^{(t)}_g, D_m$):}

\item     \hspace*{\algorithmicindent} initialize local surrogate model: ${\theta}_m^{(t+1)}=\theta_g^{(t)}$
\item     \hspace*{\algorithmicindent} \textbf{for } $t_l=1,2,...,T_l$ \textbf{do}

\item     \hspace*{\algorithmicindent} \quad update ${\theta}_m^{(t+1)}$ on $D_m$ by SGD

\item     \hspace*{\algorithmicindent} \textbf{return} ${\theta}_m^{(t+1)}$
\end{minipage}
}
\end{algorithmic}
\end{algorithm}

\subsection{Overview}

In this work, we introduce a novel \underline{FL} training approach grounded in \underline{D}ataset distillation with deep \underline{G}enerative \underline{M}odels (\alg{}). Diverging from conventional aggregation based FL training techniques like FedAvg~\cite{mcmahan2017communication}, \alg{} distinguishes itself by enabling collaborative training of a large global model with enhanced performance while maintaining a comparatively lower computational burden. In contrast to data distillation based FL methods, exemplified by FedDM~\cite{xiong2022feddm}, \alg{} sets itself apart by addressing privacy concerns more effectively than FedDM. This is achieved by exclusively transferring model parameters rather than synthetic data, ensuring a heightened level of privacy preservation.

\begin{figure}[ht]
  \begin{minipage}{\columnwidth}
    \centering
       \includegraphics[width=.4\pdfpagewidth, height=.4\pdfpagewidth]{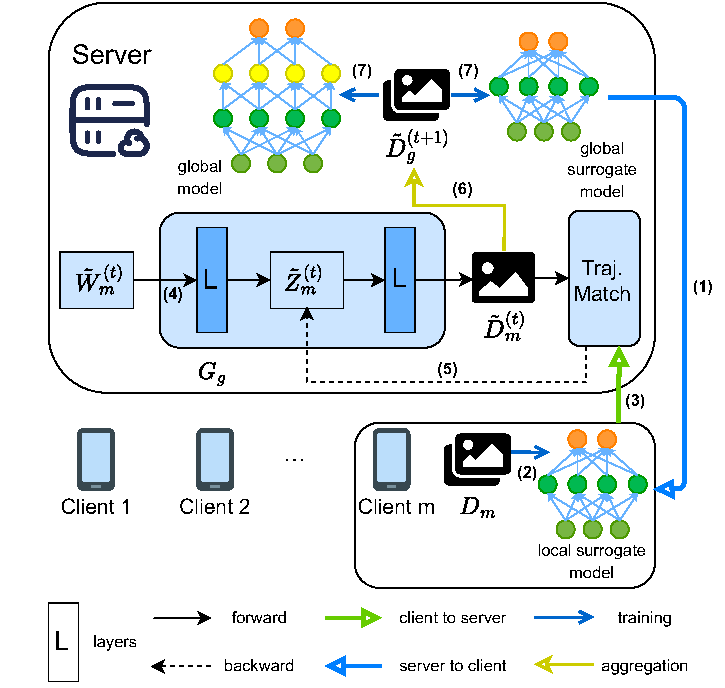}
    \caption{ An overview of \alg{}. At each global communication round $t$, (1) the server sends global surrogate model to clients, (2) clients train local surrogate models on local data, (3) clients send local surrogate models to the server, (4) the server initializes latent vectors and generates synthetic data for each client $m$, (5) the server uses MTT to update latent vectors, (6) after distilling data, the server aggregates all synthetic data and gets the synthetic dataset $\tilde{D}_g^{(t+1)}$, and (7) the server uses the synthetic dataset to train the global model and global surrogate model.}
    \label{fig:overview}
  \end{minipage}\\[1em]
  \vspace{-6mm}
\end{figure}

In particular, to reduce the computational costs for clients, we enable clients to utilize a small \emph{surrogate model}, which typically has fewer model parameters than the global model. Clients use this surrogate model for local training and send their trained models to the server. To avoid additional communication costs and privacy concerns, when the server receives the surrogate models from clients, it employs the Matching Training Trajectories (MTT) method~\cite{Cazenavette-DD-2022-trajectory} to distill synthetic data for each client. To enhance the training effectiveness of the synthetic data on the global model and the performance of \alg{} on high-resolution datasets, we leverage prior knowledge from pre-trained deep generative models to distill data. Subsequently, the server trains the global model on the distilled data. Moreover, the server uses the distilled data to train an extra global surrogate model, and then sends it back to the clients, enabling them to perform their local training in the next communication round. Fig.~\ref{fig:overview} is an overview of \alg{}, and Algorithm~\ref{alg:fedgan} summarizes our \alg{}. 

\subsection{Problem Formulation}
We begin by introducing some notations in \alg{}. Assuming that each client $m$ has its own local training dataset $D_m$ and a surrogate model parameterized by $\theta_m^{(0)}$. In each global communication round $t$, client $m$ addresses the following optimization problem:

\begin{equation}
    \begin{aligned}
        \theta_m^{(t+1)}=\min_{\theta\in B_r(\theta_g^{(t)})} \mathcal{L}(D_m;\theta),
    \end{aligned}
\end{equation}
where $\theta_g^{(t)}$ is the global surrogate model parameters sent from the server, and $B_r(\theta_g^{(t)})$ is a $r$-radius ball around $\theta_g^{(t)}$. After client $m$ sends its local surrogate model parameters $\theta_m^{(t+1)}$ to the server, the server distills synthetic data $\tilde{D}_m^{(t)}$ based on $\theta_m^{(t+1)}$. The server gets the synthetic dataset $\tilde{D}_g^{(t+1)}$ by aggregating all $\tilde{D}_m^{(t+1)}$, and trains the global model parameterized by $w_{g}^{(t)}$, together with the global surrogate model parameterized by $\theta_{g}^{(t)}$, on $\tilde{D}_g^{(t+1)}$. Subsequently, the server sends the global surrogate model parameters back to clients. In the next section, we will delve into the core component of \alg{}, which is the data distillation on the server.

\subsection{Global Data Distillation}
Our data distillation incorporates two key techniques to address different issues: distillation via Matching Training Trajectories and optimization via deep generative latents. \vspace{-3mm}

\subsubsection{Distillation via Matching Training Trajectories}
Previous FL training methods that incorporate data distillation directly transmit synthetic data between the server and clients~\cite{xiong2022feddm}, leading to privacy concerns. In contrast, \alg{} adopts a different approach where the server only receives surrogate models from clients. This prevents the server from using dataset distillation methods that require access to the actual local data of clients, such as data condensation~\cite{zhao2021dataset} and distribution matching~\cite{zhao2023dataset}. Hence, \alg{} employs an adaptive version of Matching Training Trajectories, a technique that solely utilizes checkpoints of model parameters saved throughout the training of clients' local surrogate models on their real datasets to distill data. 

Specifically, after client $m$ uploads its local surrogate model parameters $\theta_m^{(t+1)}$, the server trains a student network initialized with $\theta_g^{(t)}$ on a randomly initialized synthetic dataset $\tilde{D}_m^{(t)}$ for $T_s$ epochs, where $\theta_g^{(t)}$ is the global surrogate model parameters received by client $m$ in the last communication round. We denote the post-training student model parameters as $\hat{\theta}_g^{(t+T_s)}$, and define the training trajectory matching loss as the normalized $L_2$ distances between $\hat{\theta}_g^{(t+T_s)}$ and client $m$'s local surrogate model parameters $\theta_m^{(t+1)}$, i.e.,

\begin{equation}
    \begin{aligned}
        \mathcal{L}_{MTT}=\frac{||\hat{\theta}_g^{(t+T_s)}-{\theta}_m^{(t+1)}||^2}{||{\theta}_g^{(t)}-{\theta}_m^{(t+1)}||^2}.
    \end{aligned}
\label{equ:loss_mtt}
\end{equation}
Then the server updates the synthetic dataset $\tilde{D}_m^{(t)}$ according to the training trajectory matching loss $\mathcal{L}_{MTT}$. The server repeats the above steps $T_d$ times to obtain the updated 
synthetic dataset $\tilde{D}_m^{(t)}$.

In the above process, we directly use $\theta_g^{(t)}$ to initialize the student network instead of randomly selecting a starting epoch from client $m$'s training trajectories. This is done to reduce communication costs, allowing the server to distill data even if it only receives updated surrogate model parameters $\theta_m^{(t+1)}$ from client $m$.
Due to the server's direct use of synthetic data for global model training, as opposed to aggregating model parameters or gradients sent by clients, data distillation significantly enhances the convergence speed of training the global model, noticeably reducing the global communication rounds. More importantly, it allows the global model on the server to have a different architecture from the local surrogate models on clients' devices, thereby reducing the computational costs for clients. To further improve the generalization of synthetic data on the global model, which often has a distinct architecture from clients' local surrogate models, we employ another key technique, optimization via deep generative latents, as introduced in the next section.

\subsubsection{Optimization via Deep Generative Latents}\label{sec:distill_deep_model}

In the previous section, unlike the methods for distilling data under centralized learning and other FL training methods that distill data on the clients' side using real data for initializing synthetic data, we randomly initialize synthetic data for each global communication round. This is due to the server's inability to access the local training data of clients. However, distilling data directly using randomly initialized synthetic data tends to result in low-quality synthetic data, subsequently leading to poor performance of the global model. Furthermore, the global model and surrogate model often have different architectures, and data distillation employs the surrogate model, resulting in synthetic data that does not generalize well on the global model. Therefore, during data distillation on the server, we employ a novel technique called optimization via deep generative latents, where the server leverages prior knowledge obtained from a pre-trained generative model to optimize the distillation process.

In our \alg{}, the server possesses a pre-trained generative model denoted as $G_g$. Optimizing distilled data in the latent space of a generative model can result in better generalization on models with different architectures compared to directly optimizing in pixel space~\cite{DD-cazenavette2023generalizing}. Hence,
in each global communication round $t$, instead of using randomly initialized synthetic data, the server begins by generating latent vectors $\tilde{Z}_m^{(t)}$ for each client $m$ through some initialization methods. These latent vectors are then employed with the help of $G_g$ to generate the distilled data $\tilde{D}_m^{(t)}$. Because $G_g$ can propagate gradients, it doesn't interfere with the use of $\mathcal{L}_{MTT}$ for optimizing the latent vectors $\tilde{Z}_m^{(t)}$.

We utilized the pre-trained StyleGAN-XL as our generative model. Since StyleGAN-XL has multiple latent spaces in its synthesis network, the choice of which latent space to use for distilling synthetic data is an intriguing question. We denote the distillation space corresponding to the $n$-th intermediate layer of $G_g$ as F$n$ space. A smaller $n$ implies that distillation relies more on the prior knowledge of the generative model, resulting in more realistic synthetic data. Conversely, a larger $n$ makes the synthetic data more expressive. In our experiments, we investigate the impact of different latent spaces on both the quality of the distilled data and the training performance.

We can also leverage prior knowledge from StyleGAN to initialize latent vectors, thus improving the quality of the synthetic data. StyleGAN incorporates a mapping network, which is typically a multi-layer perceptron, responsible for mapping the input vector $\tilde{W}$ to an intermediate latent space. The mapping network transforms the latent space into a more expressive style space, offering fine-grained control over styles. Therefore, we can randomly generate some vectors $\tilde{W}_m^{(t)}$ for client $m$ and use $G_g$'s mapping network, along with the earlier layers of the synthesis network, to initialize latent vectors  $\tilde{Z}_m^{(t)}$ that possess prior knowledge.

\section{Theoretical Analysis}\label{theory:main}

In this section, we articulate the argument that, from an asymptotic standpoint, the described procedure becomes equivalent to the theoretical ideal of centralized training on the cumulative heterogeneous dataset. The central server is performing GD with gradually changing data, with iteration,
\begin{align}
    \theta^{t+1}_g = \theta^t_g-s_g\sum\limits_{m\in S_t} \nabla f(\theta^t_g;\tilde{D}^t_m) \notag
\end{align}
\noindent where $s_g$ is the step-size and $f$ is the loss function. 

Recall that the dataset distillation optimization problem for each agent is defined by,
\[
\begin{array}{rl}
\min\limits_{\tilde{D}^t_m} & f(\theta^*(\tilde{D}^t_m),D^t_m) \\
\text{s.t. } & \theta^*(\tilde{D}^t_m) \in \arg\min f(\theta,\tilde{D}^t_m)
\end{array}
\]
One can consider the federated DD component as an iterative optimization process for this particular problem, for simplicity writing each major iteration as a full gradient update,
\[
\tilde{D}^{t+1}_m = \tilde{D}^t_m - s_D \nabla_{\theta} f(\theta^*(\tilde{D}^t_m),D^t_m)\frac{d \theta^*(\tilde{D}^t_m)}{d D^t_m}
\]
where the term $\frac{d \theta^*(\tilde{D}^t_m)}{d D^t_m}$ represents the sensitivity of the training solution with respect to the distilled dataset. Consider a Stochastic Differential Equation (SDE) model of the training:\vspace{-8mm}

\begin{align}
    d\theta^t_g = -\frac{1}{|S_t|}\sum\limits_{m\in S_t} \nabla f(\theta^t_g;\tilde{D}^t_m) dt+dW_t, \notag
\end{align}
\vspace{-8mm}

\begin{align}
d\tilde{D}_t = -\nabla_{\theta} f(\theta^*(\tilde{D}^t_m),D^t_m)\frac{d\theta^*(\tilde{D}^t_m)}{d \tilde{D}^t_m}+d\tilde{W}_t \notag
\end{align}
Note that the second does not depend on $\theta_g$, it thus drifts independently. This is of course a simplification as the
Local SGD iterations for each client are constrained to be near $\theta_g$. As we do not know how far the local iterations are from optimality, and the ultimate intention is to perform dataset distillation as in the solution of the bilevel problem, we study the results with idealistic formalism.

We would need to prove some conditions for this to be the case, but the stationary distribution of this second diffusion can be written by taking the antiderivative of the drift term with respect to $\tilde{D}^t$ and taking the negative exponential
\[
\pi(\tilde{D}^t_m \vert \theta^*) \propto \exp\left\{- \beta_D f(\theta^*(\tilde{D}_m^t),D^t_m)\right\}
\]
where $\beta=1/s$, the step-size of the SGD, which also corresponds to the entropy. As $\beta\to \infty$, the distribution converges to a delta distribution on interpolating the sample, and as $\beta\to 0$, far from mean samples are more likely to be taken. Now, $\theta^*(\tilde{D}^t_m)$ also exhibits a distribution, considering the stationary Gibbs for the neural network,
\[
\pi(\theta^*\vert \tilde{D}_m^t) \propto \exp\left\{-\beta_* f(\theta,\tilde{D}_m^t)\right\}.
\]
As this network is overparametrized relative to the data, this distribution has multiple modes and a connected zero loss region in $\theta$ space. If we consider iteratively sampling in an alternating fashion from $\pi(\tilde{D}^t_m \vert \theta^*) $ to
$\pi(\theta^*\vert \tilde{D}_m^t)$, then we know from~\cite[Theorem 1]{schervish1992convergence} that the stationary distributions of $\tilde{D}^t_m$ and $\theta^*$ converge to their marginals. Let us now study the properties of the limiting marginal distributions. Detailed balance (as required for ergodicity, or convergence in distribution, see e.g.~\cite{stroock2013introduction}) requires that,
\[
\begin{array}{l}
\pi(\theta^*)\exp\left\{- \beta_D f(\theta^*,D^t_m)\right\}= \pi(\tilde{D}^t_m)\exp\left\{-\beta_* f(\theta^*,\tilde{D}_m^t)\right\}  \end{array}
\]
Taking $\beta_*\to \infty$ we see that we have convergence in distribution of, 
\[
\pi(\tilde D_m^t)= \exp\left\{- \beta_D f(\theta^*,D^t_m)\right\} \pi(\theta^*)
\]
with,
\[
\text{supp}\left(\pi(\theta^*)\right) \subseteq \arg\min f(\theta,\tilde D_m^t)
\]
Consider taking $\beta_D\to \infty$ now, and thus, 
\[
\begin{array}{l}
\hspace{-2mm}\text{supp}\left(\pi(\tilde D_m^t)\right)\subseteq  \\ 
\left\{\tilde D_m^t:\arg\min f(\theta,\tilde D_m^t)\cap \arg\min f(\theta, D_m^t) \neq \emptyset\right\}.
\end{array} 
\]
Finally, the solution of $\theta_g$ is based on a deterministic gradient descent, i.e., 
there is no stochasticity, because the entire dataset $\tilde{D}^t_m$ is small and hence can
be loaded in memory. However, $\tilde{D}^t_m$ itself is a stochastic process. Furthermore, as the network
is overparametrized relative to the size of $\tilde{D}^t_m$, it has a submanifold of zero loss
solutions. 
\begin{equation}\label{eq:idealprob}
\theta^*_g \in \arg\min_{\theta} \sum_m f(\theta;\tilde{D}^t_m)
\end{equation}
Now, if there exists $\theta^*$ such that $\theta^*\arg\min f(\theta, D_m^t)$ for all $m$, then it also solves~\eqref{eq:idealprob}. Otherwise, SGD iterations for~\eqref{eq:idealprob} involve descent by
$\sum_m \nabla_{\theta} f(\theta;\tilde{D}^t_m)$, which is the same as a linear combination of the $m$ computations to find $\arg\min f(\theta, D_m^t)$ for all $m$. Thus we have established the equivalency.

\section{Experiment}

\subsection{Experimental Setup}
\noindent \textbf{Datasets.} We evaluate our method on the CIFAR-10 dataset and five 10-class subsets of ImageNet. Specifically, we use CIFAR-10 (32$\times$32) to evaluate the performance of \alg{} on low-resolution data and employ subsets of ImageNet (128 $\times$ 128) to evaluate performance on high-resolution data. Previous works introduced some subsets of ImageNet, such as ImageWoof (dogs)~\cite{fastaiimagenette}, ImageMeow (cats) and ImageFruit (fruits)~\cite{Cazenavette-DD-2022-trajectory}, as well as ImageFood (food) and ImageMisc (miscellaneous items)~\cite{DD-cazenavette2023generalizing}. We provide a detailed list of the categories contained within each subset of ImageNet at the end of this section.

A distinguishing feature of FL is the non-identical distribution (non-i.i.d.) nature of local training data across clients. Assuming there are $M$ clients, to simulate non-i.i.d. settings, we partition the training data by Dirichlet
distribution $\text{Dir}_M(\alpha)$~\cite{Vahidian-flis, Vahidian-flis-TAI}, where $\alpha>0$ is some parameter that controls the non-i.i.d. degree. A smaller $\alpha$ implies a higher degree of non-i.i.d. By default, we have a total of $M=10$ clients, with $\alpha$ set to 0.5. Furthermore, in Section~\ref{exp:performance}, we will also investigate the performance of \alg{} under various non-i.i.d. scenarios. \\

\noindent \textbf{Compared Methods.} 
To provide a more comprehensive understanding of the efficiency of \alg{}, we select three FL training methods based on aggregation: FedAvg~\cite{mcmahan2017communication}, FedProx~\cite{li2018federated}, and FedNova~\cite{wang_tackling_2020}, as well as a method based on data distillation, FedDM~\cite{xiong2022feddm}.\\

\noindent \textbf{FL Training Settings.} We have a total of $M=10$ clients, and for each global training round, the server selects all clients for aggregation. For the CIFAR-10 dataset, we set the training batch size to 256, and for subsets of ImageNet, the corresponding training batch size is 32. By default, each client conducts local training for $T_{l}=20$ epochs on its own local training data, and we tune $T_{l}$ within the range $[5, 10, 20, 30]$. On the server side, for each global training round, the server updates the latent vector for $T_{d}=100$ iterations, and for each distillation step, the server updates the student network for $T_s=20$ epochs. The number of images per class (\emph{IPC}) is 10. The images are distilled into F5 space (the 5th layer of StyleGAN-XL) for CIFAR-10 and F12 for subsets of ImageNet. After distilling the data, the server trains the global model for $T_{g}=1000$ epochs with a training batch size of 256 for CIFAR-10 and 32 for subsets of ImageNet. In particular, we tune the layer of StyleGAN-XL in $[0, 3, 4, 5, 6, 9]$ and {IPC} in $[1, 5, 10, 20]$ for CIFAR-10.

To ensure a fair comparison, we maintain the same local training batch size as \alg{} for baseline methods. For FedAvg, FedProx, and FedNova, we also use the same local training epochs of $T_{l}=20$ as in \alg{}. For FedDM, we adopt the same IPC=10 and train the global model for $T_g=1000$ epochs with the same training batch size, using the synthetic dataset. All the experiments are run for three times, and we report the mean validation accuracy $\pm$ 1 standard deviation for each evaluation case.\\

\noindent \textbf{Network Architectures.} To demonstrate the cross-architecture generalization of \alg{}, for data distillation-based FL methods (i.e., FedDM and \alg{}), we set the local models of clients as a 5-layer ConvNet~\cite{gidaris2018dynamic} and set the global model to be some larger models. Specifically, for CIFAR-10, we evaluate global models such as ConvNet, ResNet18~\cite{he2016deep}, VGG11~\cite{simonyan2014very}, and ViT~\cite{dosovitskiy2020image}, while for subsets of ImageNet, we consider global models like ConvNet, ResNet18, and ResNet34~\cite{he2016deep}. For the remaining baseline methods, the structure of clients' models remains consistent with the server's global model.

{
\begin{table}

\begin{center}

\caption{{ \textbf{Using data distillation for FL training significantly enhances the performance of global models in extremely non-i.i.d. scenarios.} As data heterogeneity increases, \alg{} exhibits a significant advantage over all baseline methods. Impact of different values of $\alpha$ on the CIFAR-10 dataset. Each column represents a specific 
architecture of the global model. }}

    \color{black}
    \label{tab:niid2}
    \centering
 \resizebox{\columnwidth}{!}{
\begin{tabular}{ll|llllll}
            \toprule
$\alpha$     &Algorithm & ConvNet  & ResNet18   & VGG11 & ViT  & Average\\
            \midrule
            
            &FedAvg & 70.1$_{\pm0.5}$& 62.1$_{\pm 0.7}$ &66.2$_{\pm0.6}$ &54.8$_{\pm0.2}$ &63.3$_{\pm0.5}$ \\

            &FedProx   & 70.7$_{\pm 0.3}$& 63.0$_{\pm 0.6}$ &66.2$_{\pm 0.5}$ &55.0$_{\pm 0.8}$ &63.7$_{\pm 0.5}$ \\
0.9
            &FedNova  & 70.6$_{\pm 0.7}$& 62.1$_{\pm 1.0}$ &66.6$_{\pm 0.6}$ &54.5$_{\pm 0.4}$ &63.5$_{\pm 0.7}$ \\

            &FedDM &70.2$_{\pm0.2}$ &73.4$_{\pm0.2}$ &68.7$_{\pm0.4}$ &51.6$_{\pm0.1}$ &66.0$_{\pm0.3}$ \\
            &\alg{} &70.8$_{\pm0.5}$ &73.8$_{\pm0.2}$ &70.1$_{\pm0.4}$ &55.3$_{\pm0.3}$ &67.5$_{\pm0.4}$ \\

            \midrule
            
            &FedAvg &    69.4$_{\pm0.1}$         &    61.1$_{\pm0.8} $     &   64.4$_{\pm0.7}$    &  53.5$_{\pm0.2}$   &    62.1$_{\pm0.4}$   \\
            &FedProx   &   69.5$_{\pm0.5}$     &    62.3$_{\pm0.5}$    & 65.3$_{\pm0.4}$ & 53.4$_{\pm0.3}$ &   62.6$_{\pm0.4}$  \\
            
  0.5       &FedNova  &   69.5$_{\pm 0.4}$       & 60.9$_{\pm 0.7}$ &    64.8$_{\pm 0.6}$   & 54.3$_{\pm 1.2}$           &  62.4$_{\pm 0.7}$  \\
    
            &FedDM  & 68.9$_{\pm 0.3}$ & 72.9$_{\pm 0.1}$ & 68.0$_{\pm 0.3}$ & 51.0$_{\pm 0.6}$ & 65.2$_{\pm 0.3}$ \\
            &\alg{}  & 70.8$_{\pm 0.9}$ & 73.7$_{\pm 0.5}$ & 69.8$_{\pm 0.3}$ & 55.9$_{\pm 0.7}$ & 67.5$_{\pm 0.6}$\\ 
            \midrule
            
            &FedAvg & 45.3$_{\pm 3.2}$ & 38.0$_{\pm 0.9}$ & 41.1$_{\pm 2.2}$ & 43.6$_{\pm 0.3}$ & 42.0$_{\pm 1.7}$ \\
            
            &FedProx   & 45.2$_{\pm 2.2}$ & 40.2$_{\pm 0.4}$ & 40.3$_{\pm 1.9}$ & 45.6$_{\pm 0.6}$ & 42.8$_{\pm 1.3}$ \\
0.1         &FedNova  & 42.9$_{\pm 1.7}$ & 35.8$_{\pm 0.6}$ & 35.8$_{\pm 1.7}$ & 42.1$_{\pm 0.7}$ & 39.2$_{\pm 1.2}$ \\
                
            &FedDM  &62.0$_{\pm 0.3}$ & 63.6$_{\pm 0.3}$ & 61.4$_{\pm 0.2}$ & 45.0$_{\pm 0.2}$ & 58.0$_{\pm 0.3}$ \\
            &\alg{}  & 68.9$_{\pm 0.6}$ & 71.6$_{\pm 0.2}$ & 68.1$_{\pm 0.2}$ & 54.2$_{\pm 0.7}$ & 65.7$_{\pm 0.4}$ \\ 
            \midrule
                        
            &FedAvg & 14.3$_{\pm 2.7}$ & 16.9$_{\pm 2.6}$ & 19.3$_{\pm 2.9}$ & 31.1$_{\pm 1.7}$ & 20.4$_{\pm 2.5}$ \\
            
            &FedProx   & 18.3$_{\pm 2.5}$ & 16.7$_{\pm 2.5}$ & 14.8$_{\pm 3.4}$ & 28.8$_{\pm 3.3}$ & 19.7$_{\pm 2.9}$ \\
0.01        &FedNova  & 10.1$_{\pm 0.1}$ & 16.4$_{\pm 1.8}$ & 13.3$_{\pm 2.6}$ & 13.3$_{\pm 2.4}$ & 13.3$_{\pm 1.7}$ \\
                
            &FedDM  &47.8$_{\pm 0.4}$ & 48.1$_{\pm 0.7}$ & 48.9$_{\pm 0.5}$ & 36.3$_{\pm 0.4}$ & 45.3$_{\pm 0.5}$ \\
            &\alg{}  & 66.1$_{\pm 0.7}$ & 69.5$_{\pm 0.1}$ & 66.4$_{\pm 0.3}$ & 51.5$_{\pm 0.2}$ & 63.4$_{\pm 0.3}$ \\

\midrule
        \end{tabular}
     }
\end{center}
\end{table}
}

\normalsize
\subsection{Performance and Convergence Rate}\label{exp:performance}
\noindent \textbf{Performance Across Different Datasets.} We investigate the performance of \alg{} on both low-resolution data using CIFAR-10 and high-resolution data using ImageNet subsets across various non-i.i.d. degrees (different $\alpha$ values) and architectures. As shown in Table~\ref{tab:niid2}, using data distillation for FL training significantly enhances the performance of global models in extremely non-i.i.d. scenarios (e.g., $\alpha=0.01$). More importantly, our \alg{} consistently outperforms the baseline methods on CIFAR-10 across different global model architectures. Particularly, as data heterogeneity increases (i.e., $\alpha=0.9\rightarrow 0.01$), \alg{} exhibits a significant advantage over the baseline methods. This could be attributed to the way we distill data, which compensates for the challenges in training posed by high data heterogeneity. When client $i$ lacks data in class $j$, the global model parameters received by client $i$ already contain information about data in class $j$ from other clients, which allows the server to use client $i$'s local model parameters to distill data in class $j$ with relatively good quality, as depicted in Fig.~\ref{fig:cifar_lack_label}. To facilitate a quick understanding of the key findings in these extensive tables, we have presented the average accuracy, represented by the last column in Table~\ref{tab:niid2}, in Fig.~\ref{fig:avg_iid_cifar}. It is evident that, with increased heterogeneity (i.e., smaller $\alpha$ values), the superiority of \alg{} becomes more pronounced. Unlike other baseline methods, which experience a substantial decline in performance as $\alpha$ decreases, \alg{} maintains a relatively stable level of performance. We also conducted experiments on a \texttt{high resolution dataset} i.e., one of the subsets of ImageNet, ImageFruit under different data heterogeneity, and the results are shown in Table~\ref{tab:imfruit_niid} and Fig.~\ref{fig:avg_iid_fruit}.

The reported results in Table~\ref{tab:imfruit_niid} demonstrate a significant improvement achieved by \alg{} compared to the state-of-the-art. \alg{}  led to a remarkable improvement of $15\%$, and $4\%$ over FedDM for the most ($\alpha=0.01$) and the least ($\alpha=0.9$) heterogeneous setting. These findings further reinforce the claim that \emph{greater data heterogeneity correlates with amplified performance improvement in the global model trained through our dataset distillation method \alg{} }. For a concise comprehension of the key findings within these extensive tables, we have summarized the average accuracy in the last column of Table~\ref{tab:imfruit_niid} in Fig.~\ref{fig:avg_iid_fruit}.

{
\begin{table}[htbp]

\begin{center}

\caption{{ \textbf{Using data distillation for FL training significantly enhances the performance of global models in extremely non-i.i.d. scenarios.} As data heterogeneity increases, \alg{} exhibits a significant advantage over all baseline methods. Impact of different values of $\alpha$ on a \texttt{high resolution dataset}, i.e., one of the subsets of ImageNet (ImageFruit). Each column represents a specific 
architecture of the global model. }}

    \color{black}
    \label{tab:imfruit_niid}
    \centering
 \resizebox{\columnwidth}{!}{
\begin{tabular}{ll|lllll}
            \toprule
$\alpha$     &Algorithm & ConvNet  & ResNet18   & ResNet34  & Average\\
            \midrule
            &FedAvg & 43.9$_{\pm 0.3}$& 35.1$_{\pm 0.8}$ & 41.8$_{\pm 0.4}$ & 40.3$_{\pm 0.5}$ \\

            &FedProx  & 48.5$_{\pm 0.9}$& 37.1$_{\pm 0.9}$ & 38.9$_{\pm 1.4}$ & 41.5$_{\pm 1.1}$ \\

0.9        &FedNova  & 46.8$_{\pm 0.7}$& 37.6$_{\pm 0.9}$ & 42.2$_{\pm 1.1}$ & 42.2$_{\pm 0.9}$ \\

            &FedDM  & 49.7$_{\pm 0.2}$& 51.7$_{\pm 0.5}$ & 56.7$_{\pm 0.9}$ & 52.7$_{\pm 0.5}$ \\

            &\alg{} & 54.6$_{\pm 0.9}$& 55.7$_{\pm 0.7}$ & 59.1$_{\pm 0.8}$ & 56.5$_{\pm 0.8}$ \\

            \midrule
            
            &FedAvg & 42.7$_{\pm 1.9}$& 36.2$_{\pm 1.4}$ &39.5$_{\pm 2.6}$ &39.5$_{\pm 2.0}$ \\

            &FedProx  & 43.9$_{\pm 0.2}$& 34.7$_{\pm 1.8}$ &42.4$_{\pm 3.3}$ &40.4$_{\pm 1.8}$ \\

0.5         &FedNova  & 45.7$_{\pm 1.2}$& 35.0$_{\pm 0.6}$ &39.1$_{\pm 1.0}$ &40.0$_{\pm 1.0}$ \\

            &FedDM  & 45.7$_{\pm 1.7}$& 49.7$_{\pm 0.2}$ & 53.7$_{\pm 1.5}$ &49.7$_{\pm 1.1}$ \\

            &\alg{} & 54.5$_{\pm 1.5}$& 55.1$_{\pm 0.7}$ &58.9$_{\pm 0.4}$ &56.2$_{\pm 0.8}$ \\
         
            \midrule 

            &FedAvg & 42.1$_{\pm 1.5}$& 33.4$_{\pm 2.2}$ & 38.0$_{\pm 1.5}$ & 37.8$_{\pm 1.7}$ \\

            &FedProx  & 39.8$_{\pm 1.2}$& 32.1$_{\pm 2.2}$ & 34.2$_{\pm 1.3}$ & 35.4$_{\pm 1.6}$ \\

0.1        &FedNova  & 40.6$_{\pm 1.0}$& 31.5$_{\pm 1.6}$ & 34.9$_{\pm 2.5}$ & 35.6$_{\pm 1.7}$ \\

            &FedDM  & 42.3$_{\pm 1.2}$& 45.1$_{\pm 0.7}$ & 48.7$_{\pm 0.7}$ & 45.3$_{\pm 0.8}$ \\

            &\alg{} & 51.3$_{\pm 0.6}$& 51.6$_{\pm 1.8}$ & 59.1$_{\pm 1.5}$ & 54.0$_{\pm 1.3}$ \\               
    
 \midrule

            &FedAvg & 23.7$_{\pm 0.1}$& 17.7$_{\pm 0.5}$ & 16.7$_{\pm 0.9}$ & 19.3$_{\pm 0.5}$ \\

            &FedProx  & 21.8$_{\pm 1.7}$& 16.3$_{\pm 0.3}$ & 17.7$_{\pm 0.8}$ & 18.6$_{\pm 0.9}$ \\

0.01        &FedNova  & 22.3$_{\pm 1.5}$& 20.0$_{\pm 1.1}$ &17.1$_{\pm 0.6}$ &19.8$_{\pm 1.1}$ \\

            &FedDM  & 32.9$_{\pm 1.5}$& 37.3$_{\pm 0.9}$ & 34.5$_{\pm 0.7}$ & 34.9$_{\pm 1.0}$ \\

            &\alg{} & 44.9$_{\pm 1.4}$& 48.5$_{\pm 0.7}$ & 55.1$_{\pm 0.2}$ & 49.5$_{\pm 0.8}$ \\

\midrule
        \end{tabular}
     }
\end{center}
\end{table}

\begin{figure}[ht]
  \begin{minipage}{\columnwidth}
    \centering
  \includegraphics[width=.99\columnwidth]{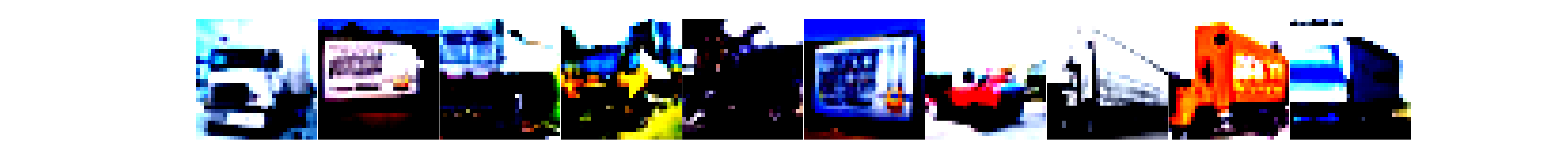}
    \caption{ Illustrations of distilled data examples labeled as ``truck''. These examples are distilled using a client's local surrogate model whose local data does not contain ``truck''.}
    \label{fig:cifar_lack_label}
  \end{minipage}\\[1em]
\end{figure}

\begin{figure}[ht]
  \begin{minipage}{\columnwidth}
    \centering
  \includegraphics[width=.99\columnwidth]{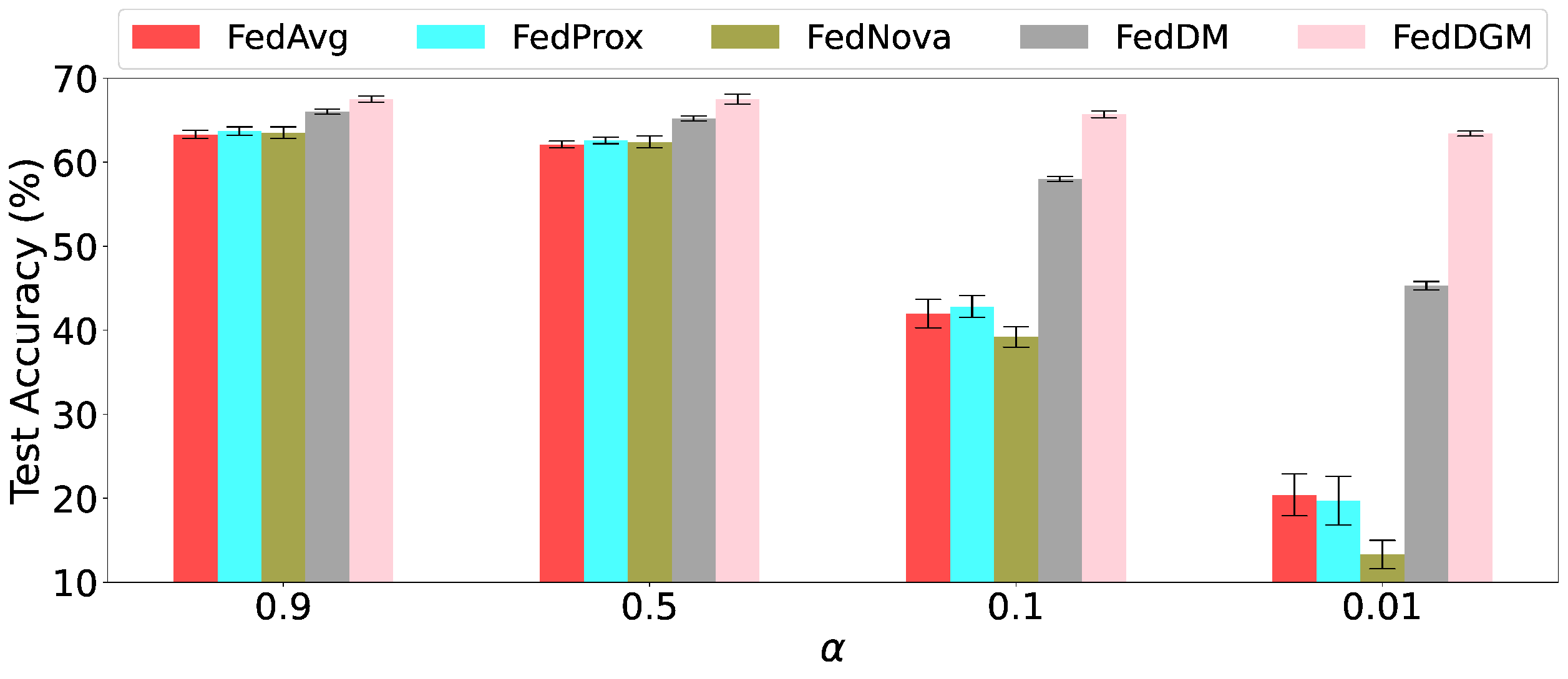}
    \caption{ \textbf{The benefit of \alg{} is particularly pronounced
when the data distributions are extremely non-i.i.d.} Impact of different $\alpha$ on the average accuracy across global models with different architectures on CIFAR10.}
    \label{fig:avg_iid_cifar}
  \end{minipage}\\[1em]
\end{figure}

\begin{figure}[ht]
  \begin{minipage}{\columnwidth}
    \centering
  \includegraphics[width=.99\columnwidth]{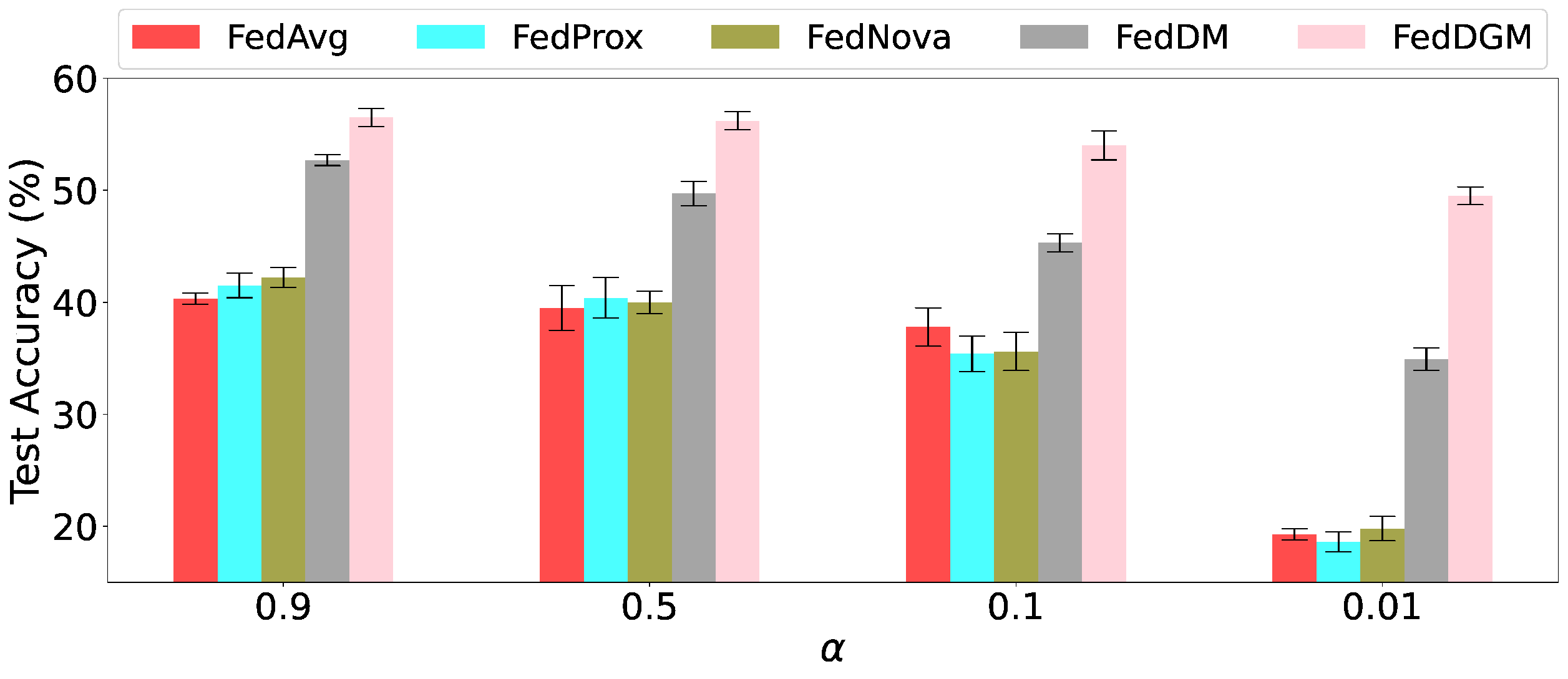}
    \caption{Impact of different $\alpha$ on the average accuracy across global models with different architectures on ImageFruit.}
    \label{fig:avg_iid_fruit}
  \end{minipage}\\[1em]
\end{figure}

In the case of high-resolution data, according to results in Table~\ref{tab:diff_subset}, utilizing data distillation for FL training significantly enhances the performance of models on high-resolution data, and \alg{} always outperforms baseline methods on all five ImageNet subsets when the server possesses global models with different architectures, such as ConvNet, ResNet18, and ResNet34. Specifically, our algorithm significantly outperforms FedDM and other aggregation-based methods on  ImageMeow, ImageWoof, and ImageFruit. For example, \alg{} achieves 57.4$\pm$0.6\% on ImageWoof when the global model is ConvNet, while the best-performing baseline method, FedDM, only achieves 47.0$\pm$1.8\%. This demonstrates a substantial improvement in our algorithm for high-resolution data. Also on ImageWoof, our algorithm achieves 72.7$\pm$0.6\% on ResNet34, exceeding the next best method by over 10\% improvement. \textit{This significant improvement highlights its strong generalization capacity across diverse model architectures. }
To facilitate ease of comparison, Fig.~\ref{fig:avg_subsets} depicts the average results across architectures, as presented in the last column of Table~\ref{tab:diff_subset}.

{
\begin{table}

\begin{center}

\caption{ \textbf{FedDGM is architecture-agnostic and performs effectively across both high-resolution and low-resolution datasets.} Test accuracy across various subsets of ImageNet. \alg{} outperforms all baseline methods across different global model architectures on these five ImageNet subsets. The default data partitioning distribution is $\text{Dir}_{10}(0.5)$.}
    \color{black}
    \label{tab:diff_subset}
    \centering
 \resizebox{\columnwidth}{!}{
\begin{tabular}{ll|llllll}
            \toprule
Dataset     &Algorithm & ConvNet  & ResNet18   & ResNet34  & Average\\
            \midrule
            &FedAvg & 48.3$_{\pm 1.5}$& 34.8$_{\pm 2.1}$ &40.7$_{\pm 1.3}$ &41.3$_{\pm 1.6}$ \\

            &FedProx  & 49.7$_{\pm 0.4}$& 38.5$_{\pm 2.0}$ &44.3$_{\pm 1.3}$ &44.2$_{\pm 1.2}$ \\

ImMeow      &FedNova  & 50.6$_{\pm 0.3}$& 34.0$_{\pm 0.6}$ &38.3$_{\pm 1.5}$ &41.0$_{\pm 0.8}$ \\

            &FedDM  & 51.7$_{\pm 1.3}$& 53.3$_{\pm 0.7}$ & 61.2$_{\pm 1.4}$ &55.4$_{\pm 1.1}$ \\

            &\alg{} & 57.8$_{\pm 1.2}$& 56.2$_{\pm 1.9}$ &65.1$_{\pm 0.6}$ &59.7$_{\pm 1.2}$ \\
            \midrule
            &FedAvg & 38.5$_{\pm 1.7}$& 25.7$_{\pm 1.5}$ &29.5$_{\pm 2.1}$ &31.2$_{\pm 1.8}$ \\

            &FedProx  & 40.5$_{\pm 1.2}$& 25.3$_{\pm 0.4}$ &29.7$_{\pm 2.0}$ &31.8$_{\pm 1.2}$ \\

ImWoof      &FedNova  & 40.7$_{\pm 1.2}$& 26.5$_{\pm 1.6}$ &27.0$_{\pm 1.6}$ &31.4$_{\pm 1.4}$ \\

            &FedDM  & 47.0$_{\pm 1.8}$& 52.7$_{\pm 0.9}$ & 60.3$_{\pm 0.2}$ &53.3$_{\pm 1.0}$ \\

            &\alg{} & 57.4$_{\pm 2.6}$& 62.3$_{\pm 1.2}$ &72.7$_{\pm 0.6}$ &64.1$_{\pm 1.5}$ \\

            \midrule
            &FedAvg & 42.7$_{\pm 1.9}$& 36.2$_{\pm 1.4}$ &39.5$_{\pm 2.6}$ &39.5$_{\pm 2.0}$ \\

            &FedProx  & 43.9$_{\pm 0.2}$& 34.7$_{\pm 1.8}$ &42.4$_{\pm 3.3}$ &40.4$_{\pm 1.8}$ \\

ImFruit      &FedNova  & 45.7$_{\pm 1.2}$& 35.0$_{\pm 0.6}$ &39.1$_{\pm 1.0}$ &40.0$_{\pm 1.0}$ \\

            &FedDM  & 45.7$_{\pm 1.7}$& 49.7$_{\pm 0.2}$ & 53.7$_{\pm 1.5}$ &49.7$_{\pm 1.1}$ \\

            &\alg{} & 54.5$_{\pm 1.5}$& 55.1$_{\pm 0.7}$ &58.9$_{\pm 0.4}$ &56.2$_{\pm 0.8}$ \\
         
            \midrule
            &FedAvg & 45.1$_{\pm 0.7}$& 32.8$_{\pm 1.3}$ &39.3$_{\pm 1.2}$ &39.1$_{\pm 1.1}$ \\

            &FedProx  & 46.3$_{\pm 2.8}$& 33.4$_{\pm 1.9}$ &36.7$_{\pm 0.7}$ &38.8$_{\pm 1.8}$ \\

ImFood      &FedNova  & 46.7$_{\pm 2.3}$& 32.7$_{\pm 1.6}$ &38.4$_{\pm 1.1}$ &39.3$_{\pm 1.7}$ \\

            &FedDM  & 50.6$_{\pm 1.4}$& 50.7$_{\pm 0.9}$ &59.8$_{\pm 0.4}$ &53.7$_{\pm 0.9}$ \\

            &\alg{} & 54.5$_{\pm 2.2}$& 52.9$_{\pm 1.3}$ &60.0$_{\pm 0.8}$ &55.8$_{\pm 1.5}$ \\

            \midrule
            &FedAvg & 56.8$_{\pm 0.9}$& 42.4$_{\pm 0.7}$ & 45.5$_{\pm 3.2}$ &48.2$_{\pm 1.6}$ \\

            &FedProx  & 59.1$_{\pm 1.9}$& 44.7$_{\pm 1.1}$ &46.1$_{\pm 1.8}$ &50.0$_{\pm 1.6}$ \\

ImMisc      &FedNova  & 58.8$_{\pm 1.0}$& 42.7$_{\pm 0.4}$ &48.3$_{\pm 0.2}$ &50.0$_{\pm 0.5}$ \\

            &FedDM  & 64.3$_{\pm 1.1}$& 63.9$_{\pm 0.6}$ & 72.2$_{\pm 0.3}$ &66.8$_{\pm 0.7}$ \\

            &\alg{} & 65.6$_{\pm 0.1}$& 66.6$_{\pm 2.1}$ &72.5$_{\pm 0.4}$ &68.2$_{\pm 1.1}$ \\
            \midrule            
        \end{tabular}
     }
\end{center}
\end{table}
}

\begin{figure}[ht]
  \begin{minipage}{\columnwidth}
    \centering
  \includegraphics[width=.99\columnwidth]{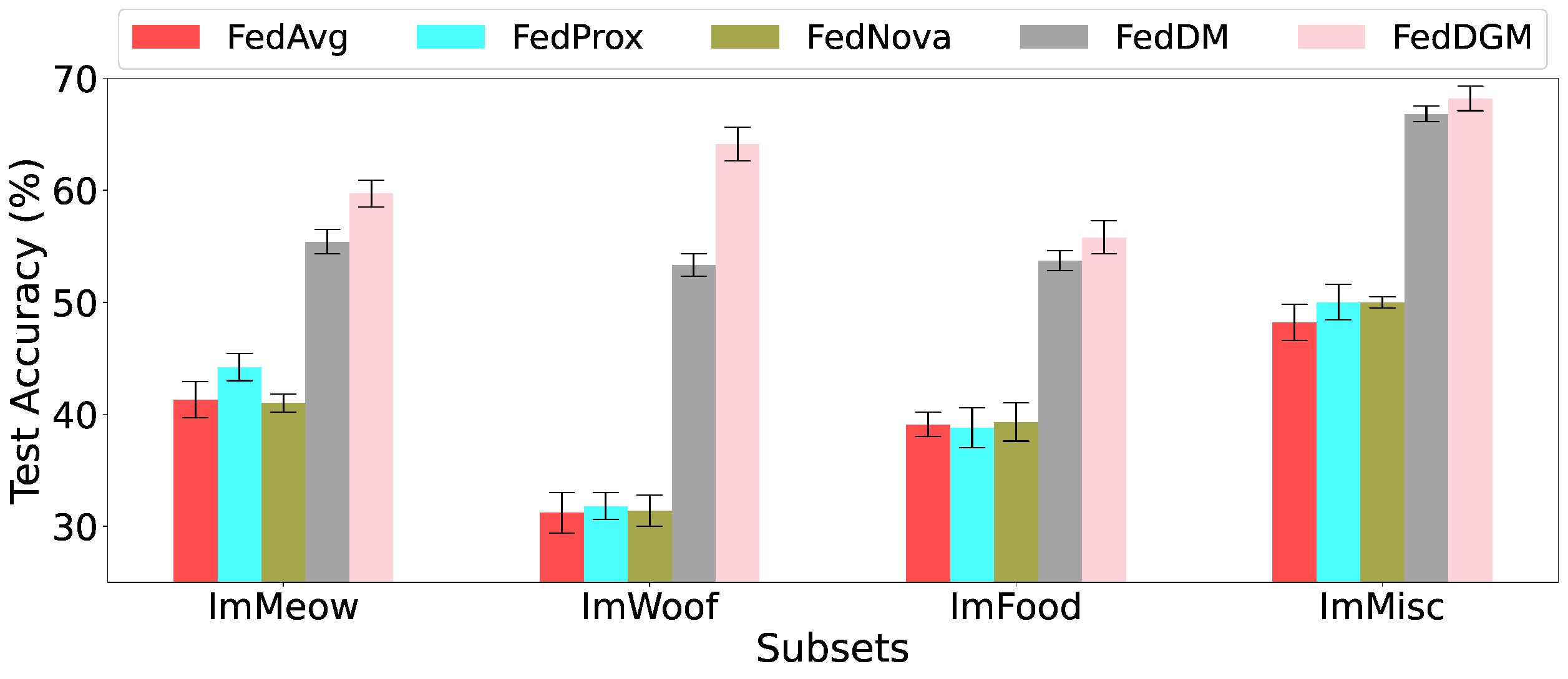}
    \caption{ \textbf{FedDGM is architecture-agnostic.} Average accuracy across global models with different architectures on various ImageNet subsets. FedDGM outperforms all baseline methods across diverse architectures.}
    \label{fig:avg_subsets}
  \end{minipage}\\[1em]
\end{figure}

\normalsize
\noindent \textbf{Convergence Rate.} Fig.~\ref{fig:cifar_main} illustrates the relationship between test accuracy and communication rounds for our \alg{} and baseline methods on CIFAR-10 across different model architectures. The data partitioning distribution is $\text{Dir}_{10}(0.5)$. Firstly, we can observe that our algorithm converges significantly faster than baseline methods on ConvNet, ResNet, and VGG11. It also converges faster than FedDM on ViT and has a comparable convergence speed with other baseline methods. In addition, we can observe that data distillation-based methods consistently exhibit a better convergence rate compared to aggregation-based methods, except when the global model is ViT. \textit{This is because the global model is trained directly on synthetic data, rather than being obtained through the aggregation of local model parameters or model updates.} The prior knowledge from deep generative models contributes to \alg{} having a better convergence rate compared to FedDM.

\begin{figure}[ht]
  \begin{minipage}{\columnwidth}
    \centering
  \includegraphics[width=.49\columnwidth]{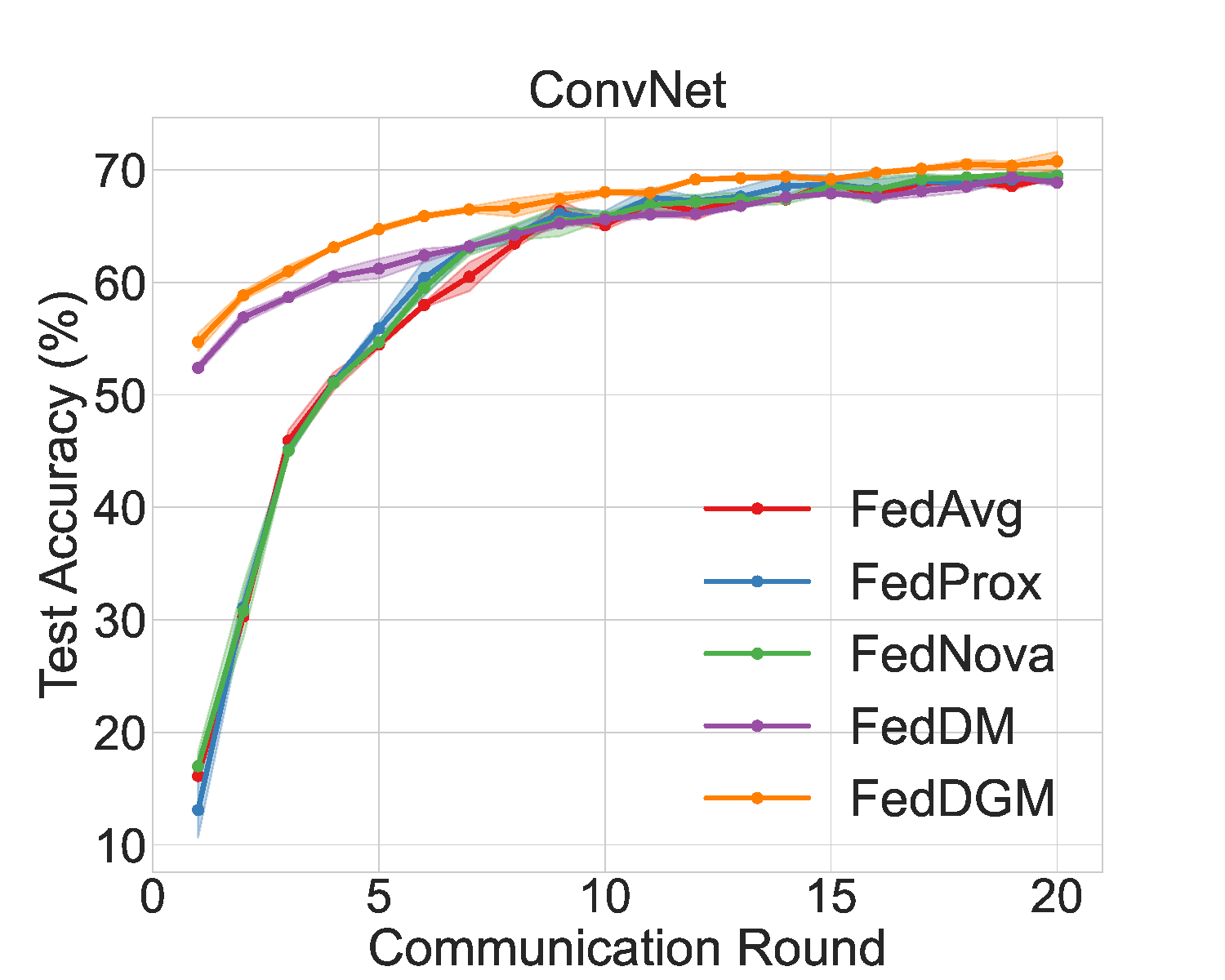}\,
  \includegraphics[width=.49\columnwidth]{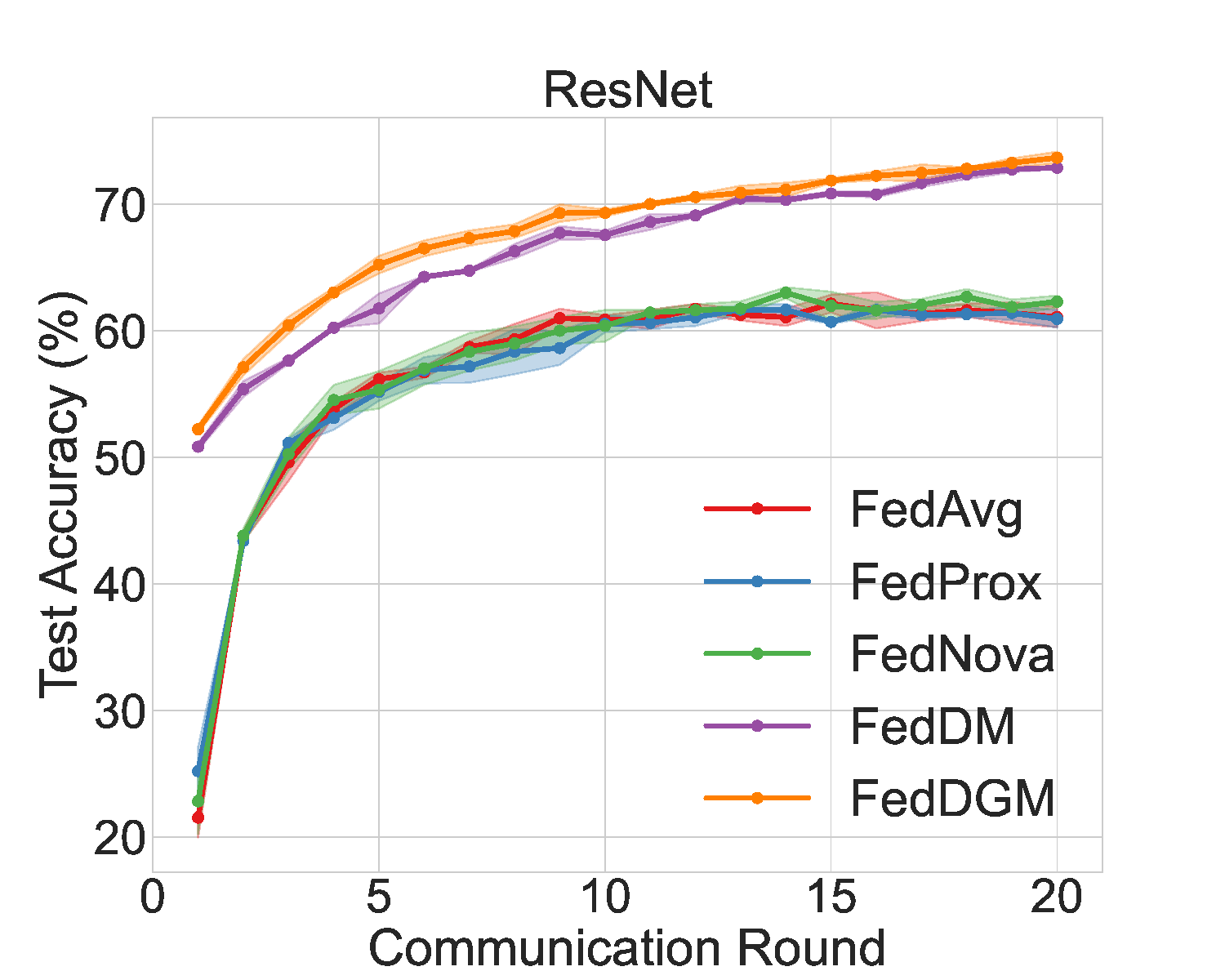}\\
  \includegraphics[width=.49\columnwidth]{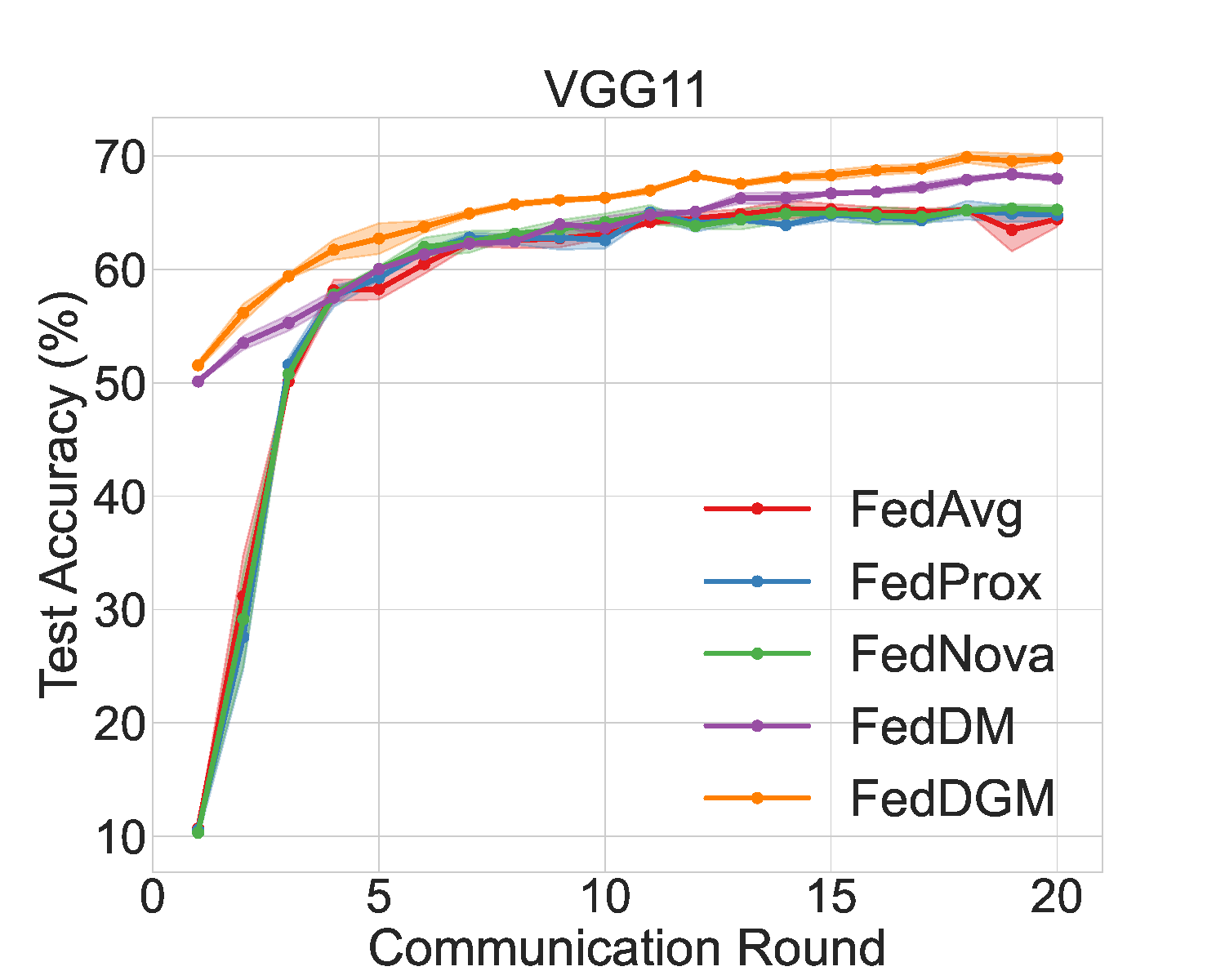}\,
  \includegraphics[width=.49\columnwidth]{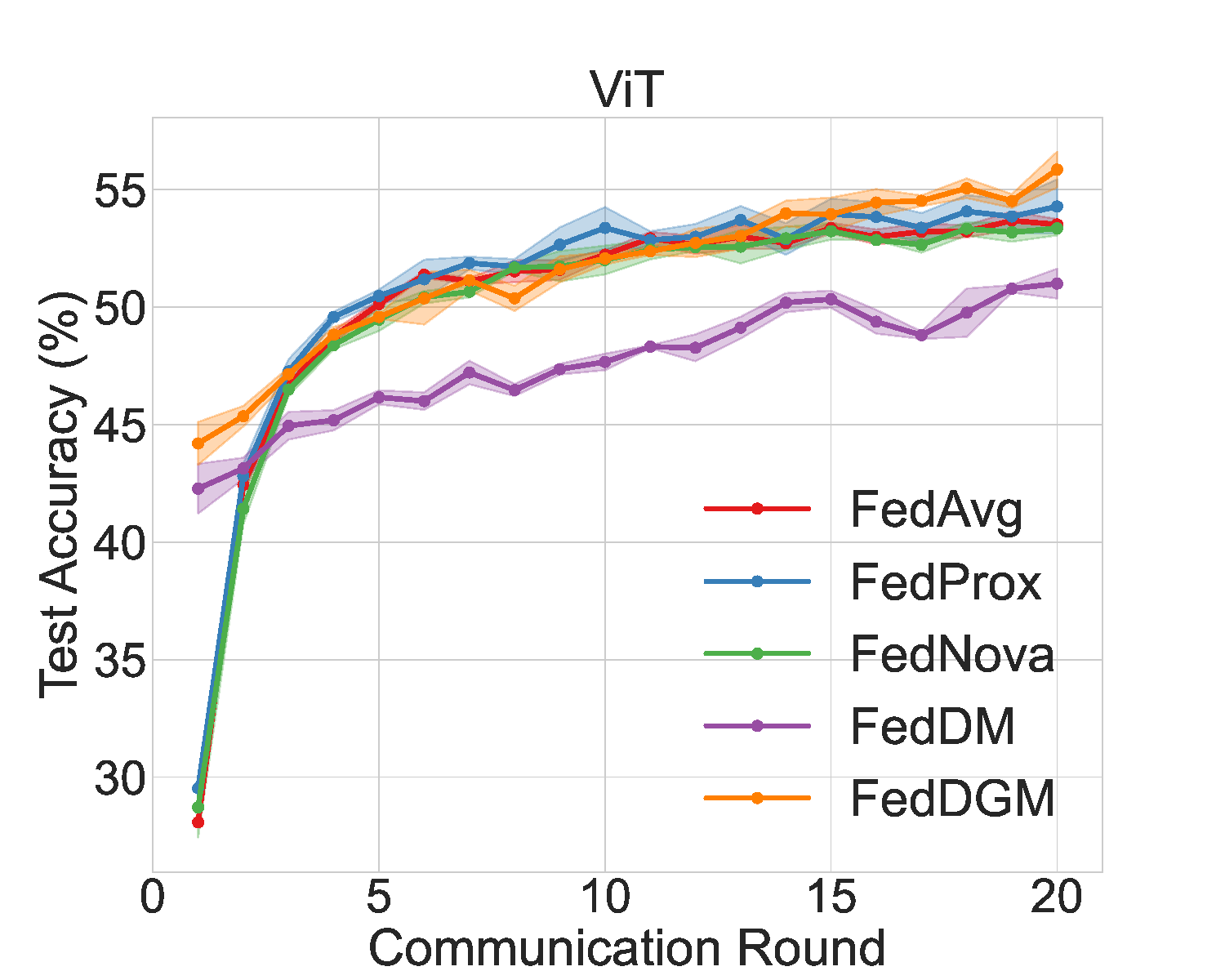}
    \caption{ \textbf{\alg{} demonstrates better performance and faster convergence rates compared to other baselines.} The relationship between test accuracy and communication rounds on CIFAR-10. }
    \label{fig:cifar_main}
  \end{minipage}\\[1em]
\end{figure}

\subsection{Impact of Hyperparameters}
\noindent \textbf{Impact of Different Latent Spaces.} As mentioned earlier in Section~\ref{sec:distill_deep_model}, data distilled from different latent spaces may exhibit noticeable differences. While keeping the data partitioning fixed, we randomly select a client and show the synthetic data belonging to the ``dog'' class generated through data distillation using six different latent spaces, as in Fig.~\ref{fig:cifar_diff_layer}. We can observe that using the latent space corresponding to an earlier layer tends to make the distilled data more realistic. To investigate which latent space results in better data generalization, we experimented with these six different latent spaces of StyleGAN-XL on CIFAR-10, and the corresponding global model performance is shown in Fig.~\ref{fig:diff_latent_space}. We can observe that using the F$5$ space results in the best generalization of distilled synthetic data across different architectures. Additionally, except for the F$9$ space, the performance difference is relatively small when parameterizing the synthetic data in other latent spaces and consistently outperforms the baseline methods.

\begin{figure}[ht]
  \begin{minipage}{\columnwidth}
    \centering
 \includegraphics[width=.39\pdfpagewidth, height=.24\pdfpagewidth]{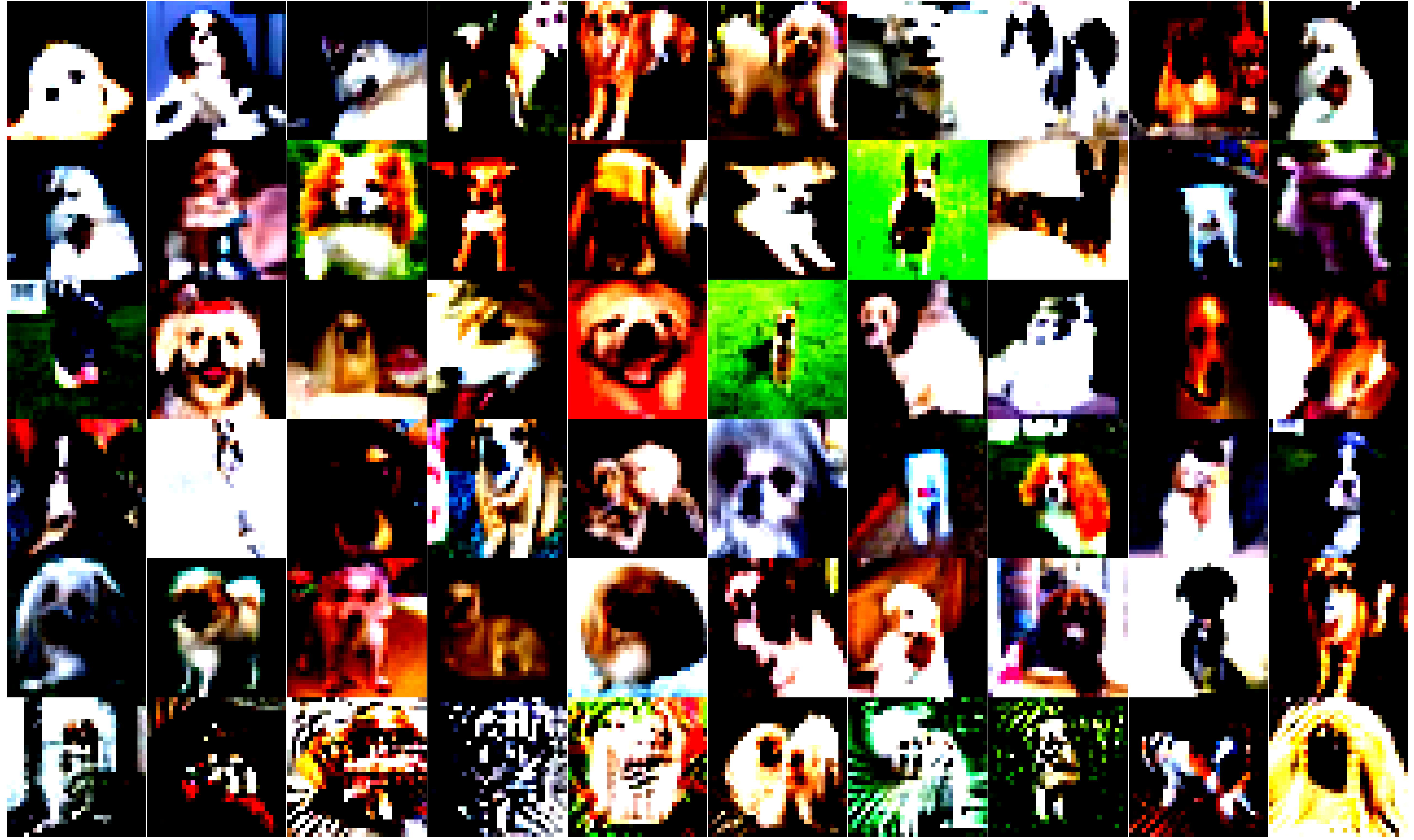}
  
    \caption{ Illustrations of distilled data generated from different latent spaces. Each row corresponds to a distinct latent space, arranged from top to bottom: F0, F3, F4, F5, F6, and F9. Employing latent spaces associated with earlier layers generally results in more realistic distilled data (i.e., those images at the top rows).}
    \label{fig:cifar_diff_layer}
  \end{minipage}\\[1em]
\end{figure}

\begin{figure}[ht]
  \begin{minipage}{\columnwidth}
    \centering
  \includegraphics[width=.99\columnwidth]{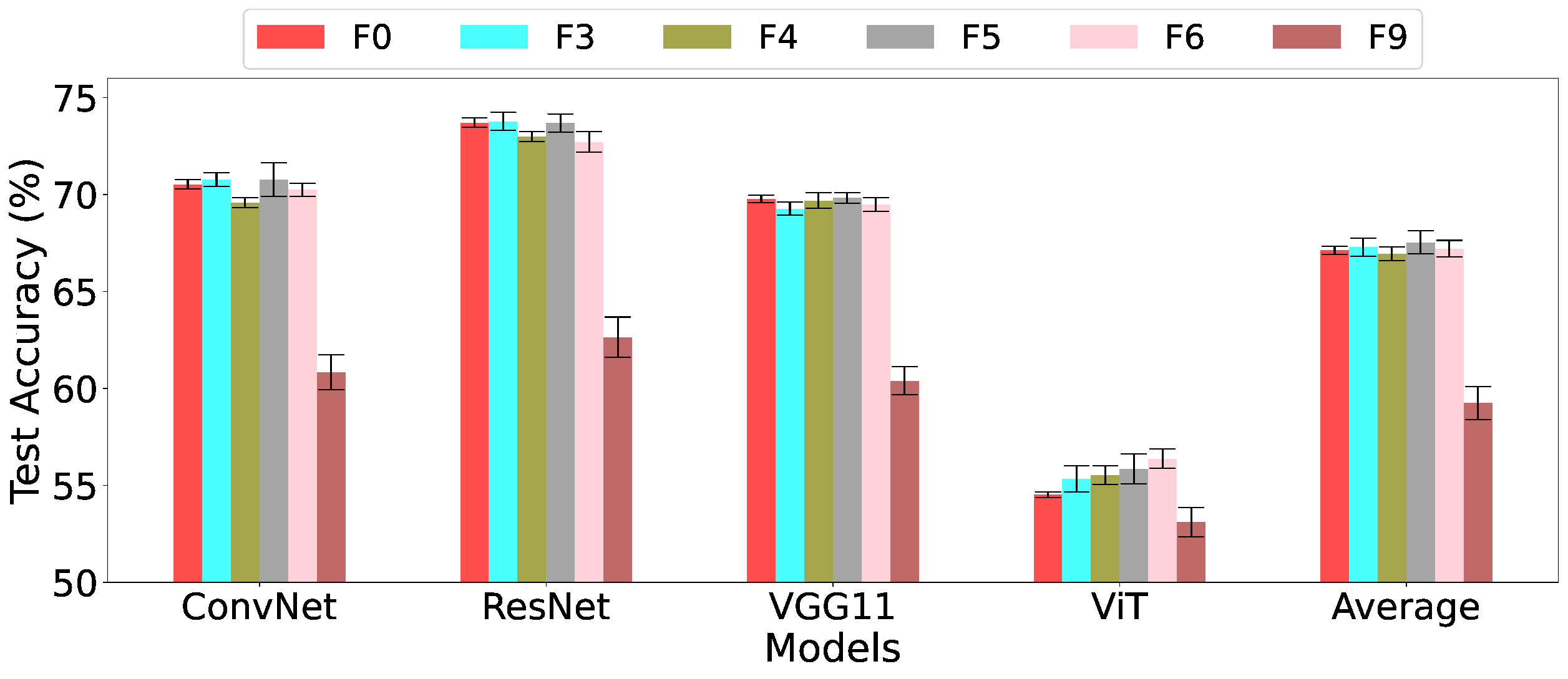}
    \caption{  Impact of distilling data in different latent spaces. Except for the F$9$ space, the performance difference among the other latent spaces is minimal. By default, we utilize the F$5$ space.}
    \label{fig:diff_latent_space}
  \end{minipage}\\[1em]
\end{figure}

\noindent \textbf{Impact of Different IPCs.} To analyze the impact of different IPC values on \alg{}, we conduct experiments to test the performance of different global model architectures with IPC values of 1, 5, 10, and 20 on CIFAR-10, and the results are presented in Fig.~\ref{fig:diff_ipc}. We can observe that as the IPC value increases, the test accuracy of global models with different architectures significantly improves. In particular, when IPC is set to 5, the performance of \alg{} is comparable to aggregation-based baseline methods. When IPC is greater than or equal to 10, \alg{} outperforms all baseline methods. Considering the trade-off between global model performance and the computational cost on the server, we set the IPC value to 10.

\begin{figure}[ht]
  \begin{minipage}{\columnwidth}
    \centering
  \includegraphics[width=.99\columnwidth]{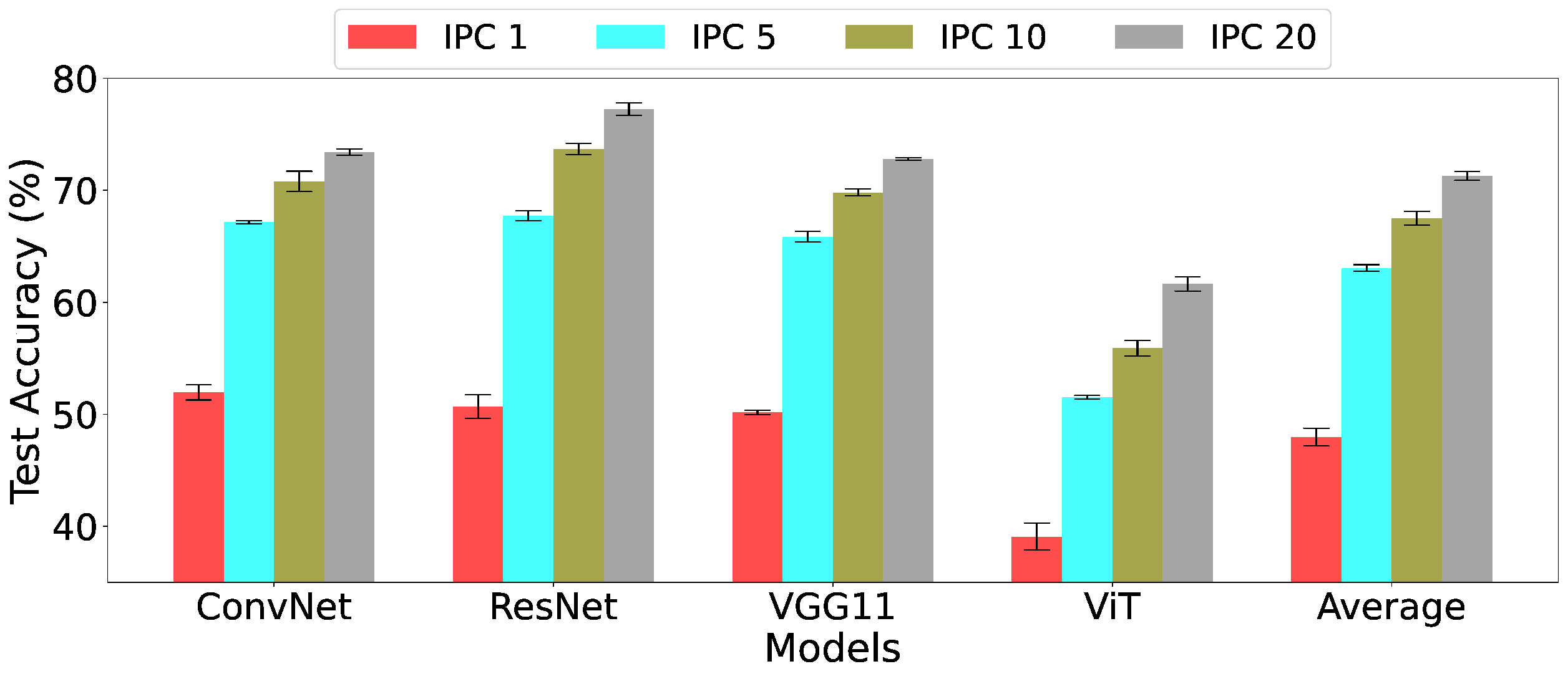}
    \caption{ \textbf{As the value of IPC increases, the test accuracy across different global model architectures significantly improves.} Impact of different IPCs.  By default, we set IPC to 10.}
    \label{fig:diff_ipc}
  \end{minipage}\\[1em]
\end{figure}

\noindent \textbf{Impact of Different Architectures of the Local Surrogate Model.} To demonstrate consistent performance of \alg{} when clients' local surrogate models have different architectures, we select a ConvNet with 2, 3, and 4 layers as the local surrogate model, respectively, and the results are depicted in Fig.~\ref{fig:diff_cnnsz}. We observe that \alg{} consistently outperforms FedDM across different architectures. Particularly, the cross-architecture performance of \alg{} remains nearly unaffected by the local surrogate model's architecture. As the architecture of the local surrogate model becomes simpler, the performance of FedDM deteriorates. Notably, when the local surrogate model is only a 2-layer ConvNet, \alg{}'s test accuracy on different global models significantly surpasses FedDM. Furthermore, using local surrogate models of a smaller size can further reduce computational costs on both the server and client sides, and significantly decreases communication costs compared to aggregation-based baseline FL methods.

\begin{figure}[ht]
 \centering
 \subfloat[2 Layers\label{fig:diff_cnnsz2}]{\includegraphics[width=0.33 \columnwidth]{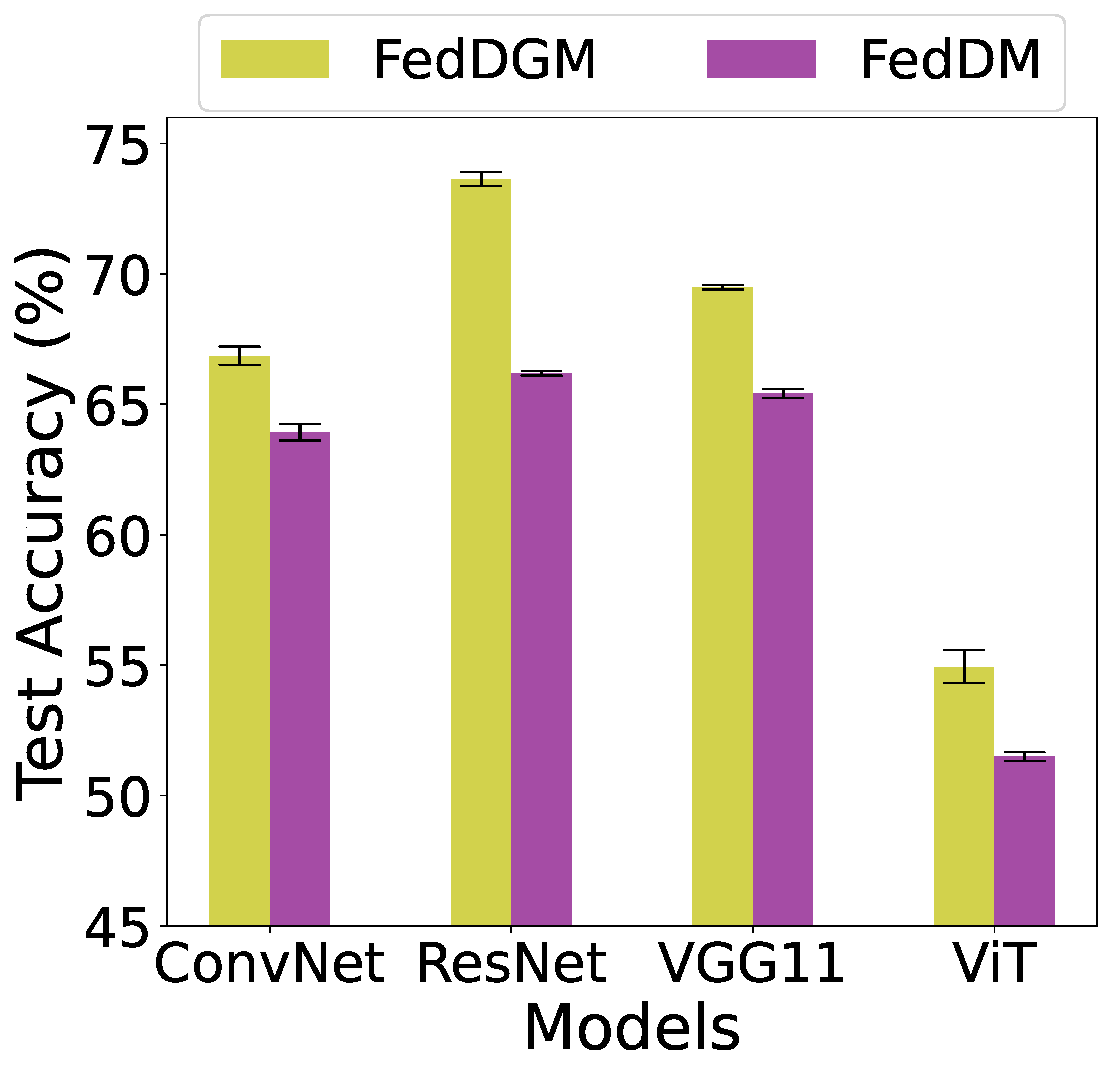}}
 \subfloat[3 Layers\label{fig:diff_cnnsz3}]{\includegraphics[width=0.33 \columnwidth]{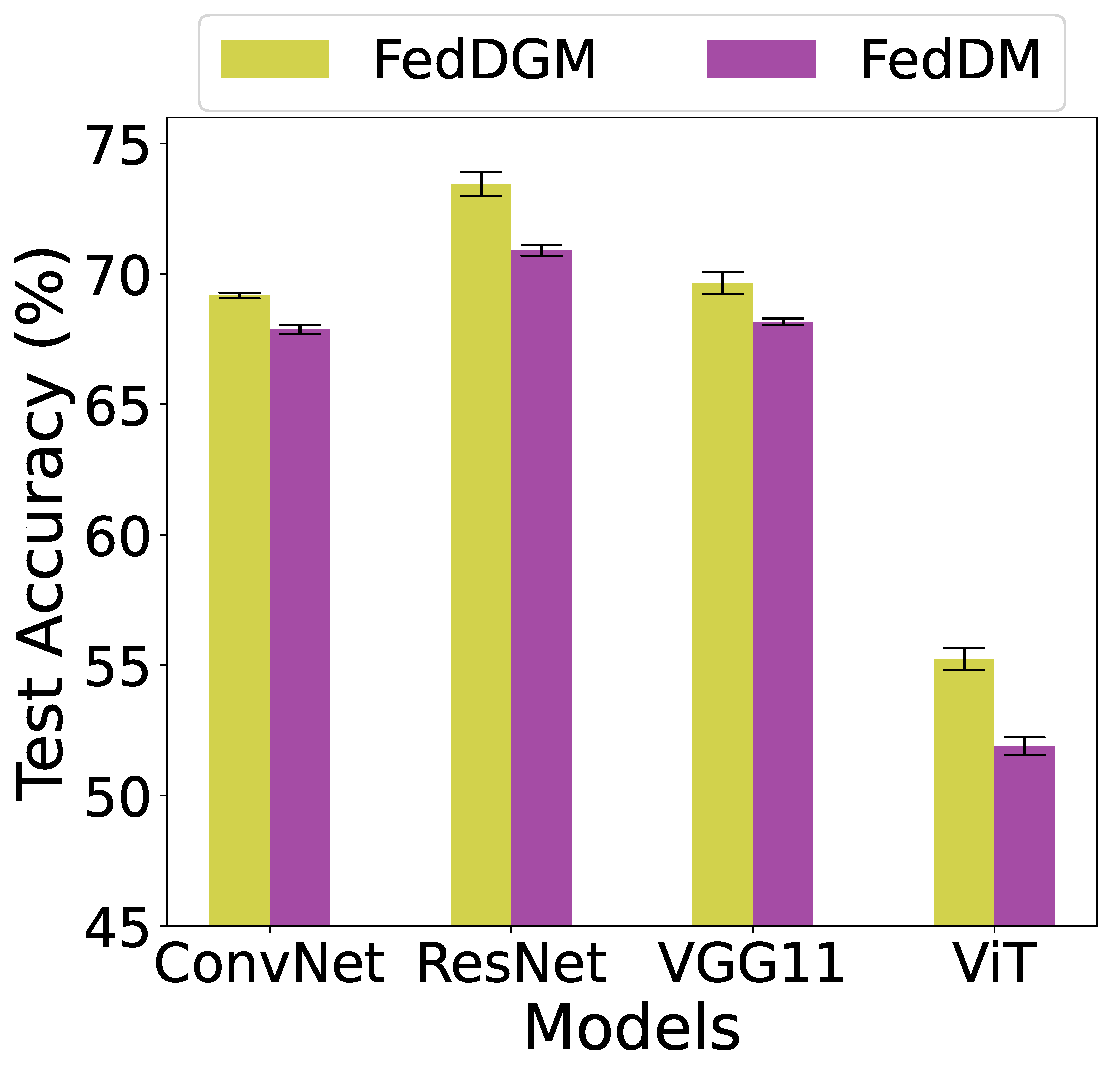}}
 \subfloat[4 Layers\label{fig:diff_cnnsz4}]{\includegraphics[width=0.33 \columnwidth]{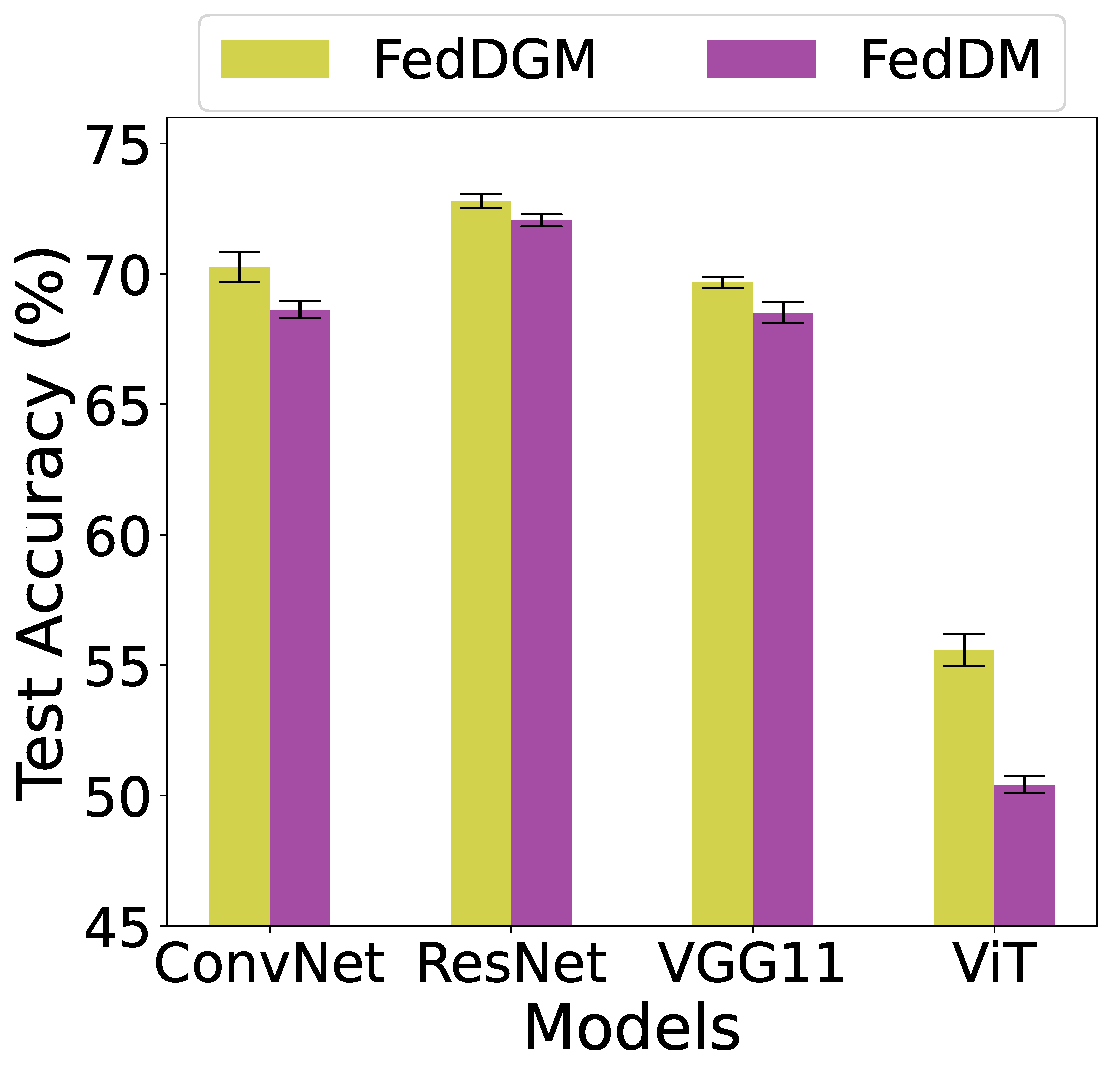}}
 \caption{ \textbf{\alg{} consistently outperforms its main competitor, FedDM across various local surrogate model architectures.} Impact of different architectures of the local surrogate model. We choose a ConvNet with 2 layers, 3 layers, and 4 layers, respectively. 
}
 \label{fig:diff_cnnsz}
\end{figure}

\noindent \textbf{Impact of Different Local Training Epochs.} 
 Fig.~\ref{fig:diff_localepoch} shows the results when clients train local surrogate models with different numbers of local epochs $T_l$ using their local real data. Interestingly, we can observe that as the number of local epochs increases, the mean accuracy of global models does not consistently improve. The global model's performance is best when the number of local epochs is set to 20, and deviating from this value, either higher or lower, results in a relatively small decrease in performance.

When the number of local epochs is small, local surrogate models are updated less frequently, leading to a slower convergence rate in the overall FL training process. On the other hand, when the number of local epochs increases, it introduces a larger gap between the post-training local surrogate model parameters $\theta_m^{(t+1)}$ and the global model parameters $\theta_g^{(t)}$ from the previous communication round. This increased gap makes it more challenging for the server to apply matching training trajectories to distill data. As a result, when the distillation step $T_d$ remains constant, a larger number of local epochs can actually lead to a decrease in the quality of the distilled data.

\begin{figure}[ht]
  \begin{minipage}{\columnwidth}
    \centering
  \includegraphics[width=.99\columnwidth]{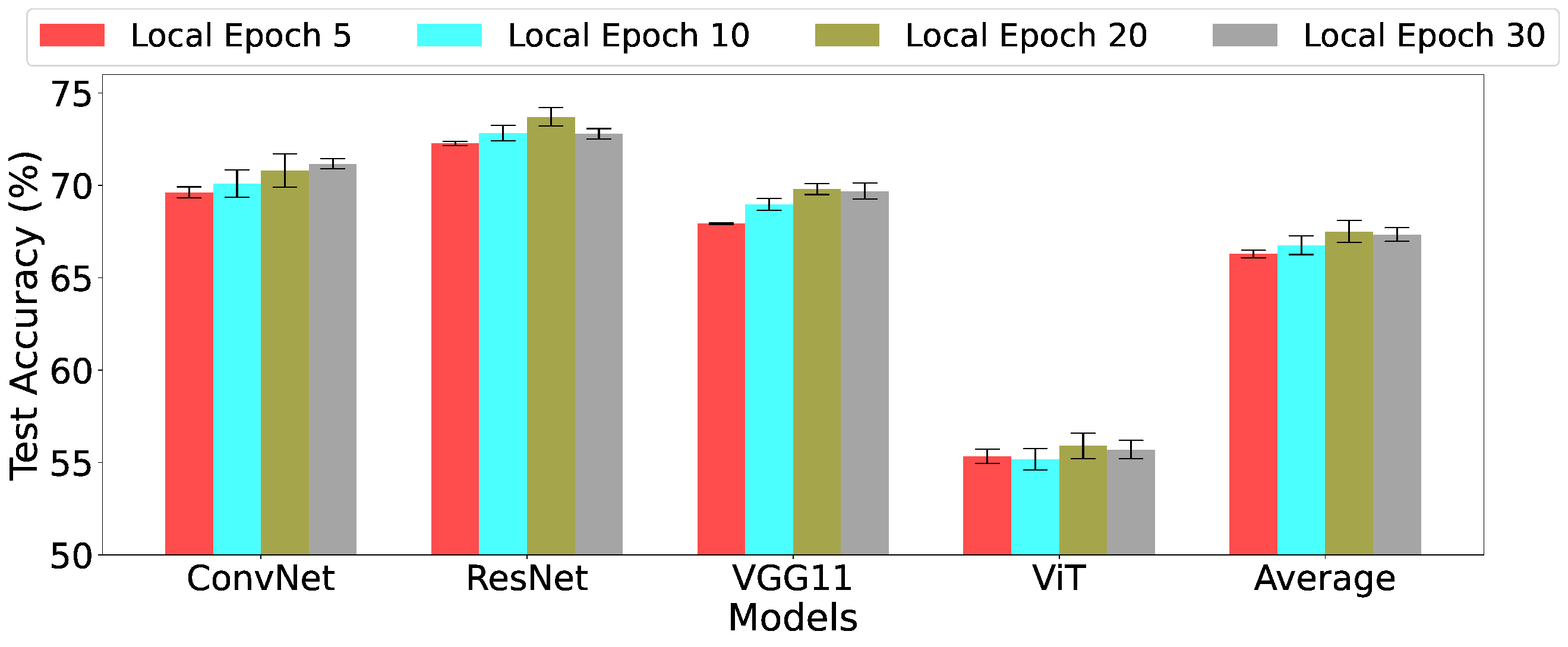}
    \caption{ \textbf{Test accuracy experiences a slight decrease when utilizing very high or very low local epochs.} Impact of different numbers of local epochs. When the number of local epochs is 20, \alg{} achieves its best cross-architecture performance. Deviating from this value, whether greater or lesser than 20, results in a slight decrease in test accuracy. }
    \label{fig:diff_localepoch}
  \end{minipage}\\[1em]
\end{figure}

\noindent \textbf{Dataset Specifications.}  Our subsets are derived from ImageNet ILSVRC2012. For each subset, we select 10 classes with similar features from the 1000 classes in ImageNet. The training set of each ImageNet subset is composed of all the training images belonging to these 10 classes, while the validation set consists of all the validation images from the same classes. Table~\ref{tab:cate_subset} shows the specific categories that make up each subset.

%
\begin{table*}[htbp]
\begin{center}

\caption{\small{Categories in ImageNet subsets}}

    \color{black}
    \label{tab:cate_subset}
    \centering
 \resizebox{\textwidth}{!}{

\begin{tabular}{l|cccccccccc}
\toprule
Subset  & 0                                                            & 1                                                        & 2                                                     & 3                                                     & 4                                                      & 5                                                          & 6                                                             & 7                                                        & 8                                                          & 9                                                          \\ \midrule
ImMeow  & \begin{tabular}[c]{@{}c@{}}Tabby\\ Cat\end{tabular}          & \begin{tabular}[c]{@{}c@{}}Bengal\\ Cat\end{tabular}     & \begin{tabular}[c]{@{}c@{}}Persian\\ Cat\end{tabular} & \begin{tabular}[c]{@{}c@{}}Siamese\\ Cat\end{tabular} & \begin{tabular}[c]{@{}c@{}}Egyptian\\ Cat\end{tabular} & Lion                                                       & Tiger                                                         & Jaguar                                                   & \begin{tabular}[c]{@{}c@{}}Snow\\ Leopard\end{tabular}     & Lynx                                                       \\ \midrule
ImWoof  & \begin{tabular}[c]{@{}c@{}}Australian\\ Terrier\end{tabular} & \begin{tabular}[c]{@{}c@{}}Border\\ Terrier\end{tabular} & Samoyed                                               & Beagle                                                & Shih-Tzu                                               & \begin{tabular}[c]{@{}c@{}}English\\ Foxhound\end{tabular} & \begin{tabular}[c]{@{}c@{}}Rhodesian\\ Ridgeback\end{tabular} & Dingo                                                    & \begin{tabular}[c]{@{}c@{}}Golden\\ Retriever\end{tabular} & \begin{tabular}[c]{@{}c@{}}English\\ Sheepdog\end{tabular} \\ \midrule
ImFruit & Pineapple                                                    & Banana                                                   & Strawberry                                            & Orange                                                & Lemon                                                  & Pomegranate                                                & Fig                                                           & \begin{tabular}[c]{@{}c@{}}Bell\\ Pepper\end{tabular}    & Cucumber                                                   & \begin{tabular}[c]{@{}c@{}}Green\\ Apple\end{tabular}      \\ \midrule
ImFood  & Cheeseburger                                                 & Hot-dog                                                   & Pretzel                                               & Pizza                                                 & French Loaf                                            & Ice-cream                                                  & Guacamole                                                     & Carbonara                                                & Bagel                                                      & Trifle                                                     \\ \midrule
ImMisc  & Bubble                                                       & \begin{tabular}[c]{@{}c@{}}Piggy\\ Bank\end{tabular}     & Stoplight                                             & Coil                                                  & Kimono                                                 & Cello                                                      & \begin{tabular}[c]{@{}c@{}}Combination\\ Lock\end{tabular}    & \begin{tabular}[c]{@{}c@{}}Triumphal\\ Arch\end{tabular} & Fountain                                                   & \begin{tabular}[c]{@{}c@{}}Spaghetti\\ Squash\end{tabular} \\ \midrule
\end{tabular}}
\end{center}
\end{table*}

\section{Conclusion} \label{conclusion}
 This paper introduces a server-side federated learning (FL) framework that leverages pre-trained deep generative models for efficient and privacy-enhanced training. This approach reduces computational demands on local devices, enabling smaller local models and facilitating the training of a larger global model on the server. Theoretical analysis shows an asymptotic resemblance to centralized training on a heterogeneous dataset. Empirical results demonstrate up to a 40\% accuracy improvement over non-dataset-distillation techniques in highly heterogeneous FL contexts, outperforming existing methods by 18\%. Notably, our framework achieves around a 10\% performance increase on high-resolution image datasets and exhibits faster convergence.



{
    \small
    \bibliographystyle{unsrt}
    \bibliography{custom}
}


\end{document}